\documentclass{article}
\usepackage[utf8]{inputenc}
\usepackage{cite}
\usepackage{hyperref}
\usepackage{graphicx,amsmath,amssymb}
\usepackage{multirow,comment,url}
\usepackage[margin=1.2in]{geometry}

\DeclareMathOperator*{\argmin}{argmin}
\DeclareMathOperator*{\argmax}{argmax}
\DeclareMathOperator{\LR}{\text{LR+}}
\DeclareMathOperator{\NB}{\text{NB}}
\DeclareMathOperator{\EC}{\text{EC}}
\DeclareMathOperator{\ECnaive}{\text{EC}_{\text{naive}}}
\DeclareMathOperator{\ECB}{\widetilde{\text{EC}}}
\DeclareMathOperator{\NECu}{\text{NEC}_b}

\DeclareMathOperator{\ECn}{\text{NEC}}
\DeclareMathOperator{\ECb}{\text{EC}_{\beta^2}}
\DeclareMathOperator{\NECb}{\text{NEC}_{\beta^2}}
\DeclareMathOperator{\NECbO}{\text{NEC}_{\beta^2\shorteq1}}
\DeclareMathOperator{\NECbT}{\text{NEC}_{\beta^2\shorteq2}}
\DeclareMathOperator{\MCC}{\text{MCC}}
\DeclareMathOperator{\FS}{\text{F}_\beta}
\DeclareMathOperator{\FSO}{\text{F}_{\beta\shorteq1}}

\DeclareMathOperator{\SPEC}{\text{SP@SE}}

\DeclareMathOperator{\XE}{\text{XE}}
\DeclareMathOperator{\BR}{\text{BR}}
\DeclareMathOperator{\ECE}{\text{ECE}}
\DeclareMathOperator{\ECEmc}{\text{ECEmc}}
\DeclareMathOperator{\EPSR}{\text{EPSR}}

\def\expv#1#2{\left\langle#1\right\rangle_{#2}}

\makeatletter
\newcommand{\shorteq}{%
  \settowidth{\@tempdima}{-}% Width of hyphen
  \resizebox{\@tempdima}{\height}{=}%
}
\makeatother

\DeclareMathOperator{\Pre}{\text{Precision}}
\DeclareMathOperator{\Rec}{\text{Recall}}

\newcommand{\eref}[1]{Equation~(\ref{#1})}

\title{Analysis and Comparison of Classification Metrics}

\author{Luciana Ferrer}

\date{Computer Science Institute, UBA-CONICET, Argentina}

\begin{document}

\maketitle

\begin{abstract}
A variety of different performance metrics are commonly used in the machine learning literature for the evaluation of classification systems. Some of the most common ones for measuring quality of hard decisions are standard and balanced accuracy,  standard and balanced error rate, F-beta score, and Matthews correlation coefficient (MCC). In this document, we review the definition of these and other metrics and compare them with the expected cost (EC), a metric introduced in every statistical learning course but rarely used in the machine learning literature. We show that both the standard and balanced error rates are special cases of the EC. Further, we show its relation with F-beta score and MCC and argue that EC is superior to these traditional metrics for being based on first principles from statistics, and for being more general, interpretable, and adaptable to any application scenario.

The metrics mentioned above measure the quality of hard decisions. Yet, most modern classification systems output continuous scores for the classes which we may want to evaluate directly. Metrics for measuring the quality of system scores include the area under the ROC curve, equal error rate, cross-entropy, Brier score, and Bayes EC or Bayes risk, among others. The last three metrics are special cases of a family of metrics given by the expected value of proper scoring rules (PSRs). We review the theory behind these metrics, showing that they are a principled way to measure the quality of the posterior probabilities produced by a system. Finally, we show how to use these metrics to compute a system's calibration loss and compare this metric with the widely-used expected calibration error (ECE), arguing that calibration loss based on PSRs is superior to the ECE for being more interpretable, more general, and directly applicable to the multi-class case, among other reasons.

We provide both theoretically-motivated discussions as well as examples to illustrate the behavior of the different metrics. The paper is accompanied by an open source repository\footnote{\url{https://github.com/luferrer/expected_cost}} that can be used to compute all the results in this paper.

\end{abstract}

\section{Introduction}

\begin{figure*}[ht]
% Plot created with the notebook in the expected_cost repository called notebooks/classification_metrics_paper/Metrics_over_hard_decisions.ipynb
\centering
\includegraphics[width=0.8\columnwidth]{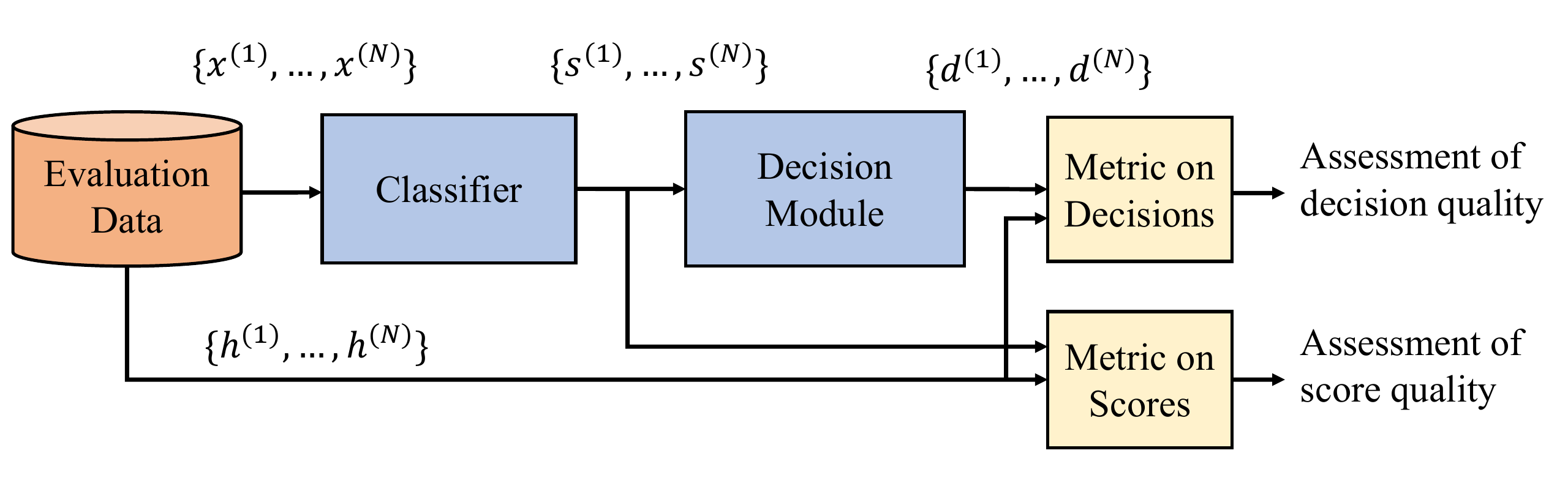}
\caption{Schematic of the pipeline for an automatic classification system composed of a score generation module, which we call classifier, and a decision module. The inputs needed to evaluate the performance of the system are shown. The evaluation data is any set of samples on which we wish to compute performance (it may even be the training data if we are curious about how the system behaves on that set). The variables $x^{(t)}$, $h^{(t)}$, $s^{(t)}$, and $d^{(t)}$, correspond to the input features, the label, the scores, and the decision for sample $t$ in the evaluation dataset. Each $s$ is a vector of dimension $K$, the number of classes, each $h$ belongs to a set $\mathcal H$  of size $K$ of possible class labels, and each $d$ belongs to a set $\mathcal D$ of size $M$ of possible decisions. As explained in Section \ref{sec:metrics_on_decisions}, $\mathcal H$ and $\mathcal D$ do not need to coincide.}
\label{fig:architecture}
\end{figure*}

Most machine learning systems have a two-stage architecture for making hard decisions \cite{introtostatslearning,bishop:ml,Hastie:statlearning}. First, a set of scores is generated for each sample. These scores are usually, though not always, designed to be  the posterior probabilities for each class given the input sample. Then, a separate stage makes the final decision based on these scores. As will be explained in more detail later, the set of possible decisions does not need to coincide with the set of true classes. It could, instead, be a set of possible actions that may be taken given a sample. Figure \ref{fig:architecture} shows a schematic of this two-stage architecture.

Development of machine learning systems requires continuous evaluation of the performance. In supervised learning, an objective function reflecting the performance of the model on the desired task (or a differentiable proxy for it) is used for training.  During development, model architecture, hyperparameters, training approaches, and many other system characteristics are determined based on the performance measured on some validation set. Finally, the performance of the resulting system is estimated on a separate test set. In each of these steps it is essential that the metric used to evaluate performance be reflective of what is important for the application of interest. In the simplest case, when all errors are equally costly, the standard error rate may be an appropriate metric for evaluation. In some applications, though, the impact of different types of error are widely different from each other and the standard error rate may not correctly reflect what is relevant in those scenarios. In other cases, it is not known ahead of time how the system will be used, what the class priors will be or which errors will be more costly. In each of these scenarios a different metric needs to be designed for the specific use case.

In this paper, we review and compare many of the standard and some non-standard metrics that can be used for evaluating the performance of a classification system.
First, we discuss metrics for assessing the quality of hard decisions. Such metrics are, in fact, completely agnostic to how the decisions are made. We describe the most standard metrics like the F-beta score \cite{vanRijsbergen79}, accuracy and error rate \cite{Hastie:statlearning}, among others, as well as a non-standard family of metrics given by the expected value of a cost function, which we call EC (expected cost). The EC is described in every statistical learning book and many other works \cite{bishop:ml,Hastie:statlearning,elkan2001,DeGroot70}, but it is unfortunately rarely used in the machine learning literature. We show that the EC is a generalization of the standard and balanced error rates (which are, in turn, one minus the standard and balanced accuracy) and argue that it is superior to many of the other metrics used in the literature, being more intuitive and flexible and having various useful properties that facilitate its interpretation. 

In Section \ref{sec:metrics_on_scores} we discuss metrics for assessing the quality of the scores. Some of those metrics assume the scores are posterior probabilities of the classes, while others do not. We describe the area under the receiver-operating-characteristic (ROC) curve, the equal error rate, and other standard metrics. Further, we describe a family of metrics obtained as the expected value of a proper scoring rule (PSR) \cite{properscoring,brummer_thesis}, which are designed to measure the performance of posterior probabilities. As such, they are sensitive to calibration issues. We discuss the calibration problem and explain how expected PSRs can be used to measure a system's calibration quality in a way that is more robust and informative than the expected calibration error (ECE) widely used in the current calibration literature \cite{guo:17}.

For each category of metrics, those that measure the quality of hard decisions and those that measure the quality of scores, we define and compare the various metrics with our recommended one for each case, and show examples to illustrate their behavior. The code used to generate all plots and tables in this paper is available at \url{https://github.com/luferrer/expected_cost}. 

\section{Metrics for Hard Decisions}
\label{sec:metrics_on_decisions}

In this section we will describe and compare classification metrics that are computed over hard decisions. We will take the set of possible true classes to be $\mathcal H = \{H_1, \ldots, H_K\}$ and the set of possible decisions made by the system to be $\mathcal D = \{D_1, \ldots, D_M\}$. Given a dataset $\{(h_1, d_1), \ldots, (h_N, d_N)\}$,  where $h^{(t)} \in \mathcal H$ and $d^{(t)} \in \mathcal D$ are the class label and the decision made by the system for sample $t$, we want to compute a metric that quantifies the quality of those decisions. 
To this end, we can compute the confusion matrix, which is composed of the number of samples, $N_{ij}$, corresponding to each combination of class and decision:
\begin{equation}
N_{ij} = \sum_{t=1}^N I(h^{(t)}=H_i)I(d^{(t)}=D_j) \label{eq:conf_matrix}
\end{equation}
where $I$ is the indicator function which is 1 when its argument is true, and 0 otherwise.
We will use $N_{i*}=\sum_j N_{ij}$ to refer to the total number of samples of class $H_i$, and $N_{*j}=\sum_i N_{ij}$ to refer to the number of samples for which the system made decision $D_j$. For example, the confusion matrix for the binary classification  case is given by:
\begin{table}[ht]
\centering
\begin{tabular}{l|l|c|c|c}
\multicolumn{2}{c}{}&\multicolumn{2}{c}{Decision}&\\
\cline{3-4}
\multicolumn{2}{c|}{}&$D_1$ & $D_2$&\multicolumn{1}{c}{Total}\\
\cline{2-4}
\multirow{2}{*}{True class}
& $H_1$ & $N_{11}$ & $N_{12}$ & $N_{1*}$ \\ \cline{2-4}
& $H_2$ & $N_{21}$ & $N_{22}$ & $N_{2*}$ \\ \cline{2-4}
\multicolumn{1}{c}{} & \multicolumn{1}{c}{Total} & \multicolumn{1}{c}{$N_{*1}$} & \multicolumn{    1}{c}{$N_{*2}$} & \multicolumn{1}{c}{$N$}\\
\end{tabular}
\end{table}

In this section, we take $\mathcal D = \mathcal H$, with one decision corresponding to each class. Hence, in the table above $D_1 = H_1$ and $D_2=H_2$. Yet, in the general case, the set of decisions may be different from the set of classes. In a medical application, for example, the classes could be ``patient has tumor", ``patient does not have a tumor". The task is to make a decision based, for example, on an MRI of the brain. Now, the decisions do not necessarily need to be one of the two classes. They could instead be the actions that the doctor could take based on the evidence (the MRI image). The actions could be, for example: ``perform surgery", ``send home", ``do more tests". The third decision would be taken in cases in which the system was not certain enough to make a decision to operate or send the patient home. Internally, the system would optimize the EC for a set of costs $c_{ij}$ to make the optimal decision for each case. In Section \ref{sec:metrics_on_scores} we will discuss how optimal decisions are made. For now, we simply assume decisions have already been made using some criteria which may or may not be optimal. 

\subsection{Expected cost}
\label{sec:ec}
The expected cost (EC) is defined as a generalization of the probability of error for cases in which errors cannot all be considered to have equally severe consequences (see, for example, Section 1.5 in \cite{bishop:ml}, Section 2.4 in \cite{Hastie:statlearning},  \cite{elkan2001}, or \cite{DeGroot70}). In the context of decision theory, the EC is used as the function to be minimized in order to make optimal decisions when the system outputs posterior probabilities. This aspect of the EC will be discussed in Section \ref{sec:metrics_on_scores}. For now, we will simply use it as a performance metric that can be computed over already-made hard decisions.  

The EC has been widely adopted by the speech processing community as an evaluation metric for the tasks of speaker verification and language recognition (see, e.g., \cite{brummer_thesis,brummer21_interspeech,van2007introduction}), where it is also the metric of choice for the periodic evaluations organized by NIST \cite{NIST_SREs}. Other than for these tasks, though, we are not aware of other applications for which the EC is used as standard practice, though some papers stand out proposing decision theory and, as a consequence, the use of EC for medical tasks \cite{ashby2000,Kornak2011}.

In this work, we will directly introduce the EC as the empirical estimate of the expected cost, i.e., as an average over the evaluation samples. To define the EC we need a \emph{cost matrix} with components $c_{ij}$ which represent the cost we believe that the system should incur for making decision $j$ when the true class of the samples was $i$. These costs are highly dependent on the application and should usually be defined in consultation with experts in the specific task of interest.

The empirical estimate of the expected cost is simply given by the average of the costs incurred over all samples in the evaluation dataset:
\begin{eqnarray}
    \EC & =&\frac{1}{N} \sum_{t=1}^N C(h^{(t)},d^{(t)}) \label{eq:ec_orig}
\end{eqnarray}
where $C(h^{(t)},d^{(t)})$ is the cost incurred when the system makes decision $d^{(t)}$ for a sample of true class $h^{(t)}$. We can rewrite this expression as:
\begin{eqnarray}
\EC    & = &\frac{1}{N}\sum_{i=1}^K \sum_{j=1}^D c_{ij} N_{ij} \label{eq:ecN}\\
    & = &\sum_{i=1}^K \sum_{j=1}^D c_{ij} P_i R_{ij}
        \label{eq:ec}
\end{eqnarray}
where:
\begin{itemize}
    \item  $c_{ij} = C(H_i,D_j)$,
    \item $R_{ij} = N_{ij}/N_{i*}$ is the fraction of samples of class $H_i$ for which the system made decision $D_j$, and
    \item $P_i = N_{i*}/N$ is the empirical estimate of the prior probability (i.e., the frequency or prevalence) of class $H_i$ in the evaluation data.
\end{itemize}
The first line in this equation is derived by using the fact that the cost is the same for each combination of class $H_i$ and decision $D_j$. Hence, we can convert the summation over the samples in \eref{eq:ec_orig} to a (double) summation over every combination of $i$ and $j$ by multiplying the corresponding cost by the number of samples that correspond to that combination of class and decision, $N_{ij}$.
Without loss of generality (see, for example, \cite{DeGroot70,brummer_thesis}), it can be assumed that all costs are positive. Hence, under this assumption, the EC is always non-negative.

The second expression for the EC, \eref{eq:ec}, allows for a very useful generalization. While $P_i$ is originally given by $N_{i*}/N$, $P_i$ can potentially be changed arbitrarily. This allows us to set the priors so that they coincide with those we expect to see when the system is deployed which, in many applications, do not necessarily coincide with those observed in the evaluation data. Rather than forcing the evaluation data to have the  priors we expect to see in practice, which would imply downsampling or upsampling the data, we can simply use those priors when computing the cost and use the evaluation data as-is to compute the $R_{ij}$ values.

\subsubsection{Normalized Expected Cost}
\label{sec:norm_ec}
The range of values that the EC, as defined in \eref{eq:ec}, can take directly depends on the costs and priors used to compute it. This makes it hard to assess whether a certain value of EC is good or bad. For this reason, it is convenient to compute a normalized version of the EC. The value used for normalization is the EC for the best naive system that does not have access to the input sample and, hence, always makes the same decision. The system is free to choose this decision to optimize the EC for the chosen costs and priors. The EC for this naive system is given by:
\begin{eqnarray}
     \ECnaive = \min_{\hat \jmath}\sum_{i=1}^K c_{i\hat \jmath} P_i \label{eq:ecn}
\end{eqnarray}
This expression can be easily derived by considering that, for a system that always makes the same decision $D_{\hat \jmath}$, $R_{ij}$ is 0 for all $j \neq \hat \jmath$ and 1 for $j=\hat \jmath$. Plugging those values of $R_{ij}$ in \eref{eq:ec} tells us that the EC for that system is $\sum_{i=1}^K c_{i\hat \jmath} P_i$. The minimum of that expression over all $\hat \jmath$ is the EC of the best naive system.
The normalized EC is then given by:
\begin{eqnarray}
    \ECn = \frac{\EC}{\ECnaive} \label{eq:ecnorm}
\end{eqnarray}

This metric has an essential property: While its value can be larger than~1, this only happens for systems that are worse than a naive system. This gives us a very useful reference: we want the $\ECn$ to be smaller than 1; if it is not, we are better off throwing away the system and outputting always the same least-costly decision. If we had access to the internals of the system instead of just its decisions, we would also be able to reduce the $\ECn$ to values below 1 by calibrating the scores from which decisions are made or changing the decision regions to optimize the $\ECn$ of interest. In the binary case, this can be achieved by simply tuning the decision threshold. A system for which decisions are optimized in this way would never have an $\ECn$ larger than 1. Hence, a system with $\ECn$ larger than 1 indicates something has gone wrong in the decision process. We will discuss issues of threshold selection and calibration in Section~ \ref{sec:metrics_on_scores}.

\subsubsection{The Special Case of Binary Classification with Square Cost Matrix}

For binary classification with the cost matrix defined such that there is one decision corresponding to each true class and the costs for correct decisions are 0, the NEC reduces to:
\begin{eqnarray}
    \ECn & =&  \frac{c_{12} P_1 R_{12} +  c_{21} P_2 R_{21}}{\min(c_{12} P_1,c_{21} P_2)} \\
    & = & \begin{cases}
    \alpha  \ R_{12} +  R_{21} & \text{if } \alpha \geq 1\\
    R_{12} +  1/\alpha  \ R_{21} & \text{otherwise} \label{eq:ec2}
\end{cases}
\end{eqnarray}
where 
\begin{eqnarray}
\alpha = \frac{c_{12} P_1}{c_{21} P_2} \label{eq:alpha}
\end{eqnarray}
Hence, we can see that, in the binary case, there is actually only one free parameter to choose in the NEC. All combinations of priors and costs that lead to the same $\alpha$ correspond to the same metric.

\subsection{Error Rate and Accuracy}
\label{sec:accuracy}

The standard error rate (also called total error or probability of error) is given by:
\begin{eqnarray}
    \mathrm{ER} & = & \frac{1}{N} \sum_{i=1}^K \sum_{j\neq i} N_{ij}
    \label{eq:acc2}
\end{eqnarray}
Comparing this equation with \eref{eq:ecN} we can see that the ER is a particular case of EC obtained when $c_{ij}=1$ for $i\neq 1$ and $c_{ij}=0$ for $i=1$, which we will call the ``0-1 cost matrix''. Further, since the accuracy is given by 1 minus the ER, we can see that accuracy is also just a trivial function of one specific case of EC.

Note that in the ER computation all samples are weighted equally. Hence, when the classes are highly imbalanced, the accuracy and ER become somewhat insensitive to the performance in the minority classes. In these cases, when errors in the detection of the minority classes are considered more severe than those in the majority classes, the balanced ER (also known as weighted ER) is used as metric instead of the standard ER. This metric is defined as the average of the ER values per class. That is:
\begin{eqnarray}
    \mathrm{BalER} & = & \frac{1}{K} \sum_{i=1}^K \frac{1}{N_{i*}} \sum_{j\neq i} N_{ij}
    \label{eq:balacc2}
\end{eqnarray}
where each term inside the sum is the ER for one specific class. This expression also coincides with an EC, when the costs are selected such that $c_{ij} = N/(K N_{i*})=1/(K P_i)$ for $i\neq j$ and $c_{ij}=0$ for $i = j$. 
As a consequence of setting the costs this way, all error rates have the same influence on the final metric. Note that by deriving the balanced error rate as a special case of the EC we can see exactly what is being assumed when we compute this metric: that the cost for each type of error is given by the inverse of the frequency of the true class of the sample. Sometimes, the balanced accuracy, given by 1 minus the balanced ER, is used as metric.

The general form of EC offers much greater flexibility than these two special cases of the standard or balanced error rates, allowing us to accommodate cases in which the errors are neither all equally costly (as in the error rate) nor as costly as the inverse of the prior for each class (as in the balanced error rate). The EC allows us to think of each type of error independently and set their cost accordingly to their severity. 

\subsection{Net Benefit}
Net benefit is a metric used in binary classification for some medical applications \cite{Vickers2016}. It is defined as:
\begin{eqnarray}
\NB & = & \frac{N_{22}}{N} - \frac{p}{1-p} \frac{N_{12}}{N}
\end{eqnarray}
where $p$ is a parameter of the metric. We can rewrite this equation as follows:
\begin{eqnarray}
\NB & = & P_2 R_{22}  - \frac{p}{1-p} P_1 R_{12}  \\
& = & P_2-(P_2 R_{21} +  \frac{p}{1-p}P_1R_{12})
\end{eqnarray}
where we used the fact that $R_{22}+R_{21}=1$. 
The term in parenthesis in this equation is the EC for binary classification when taking $c_{11}=c_{22}=0$, $c_{21}=1$, and $c_{12}=p/(1-p)$. We can then express NB as a function of that normalized EC, which we call $\ECn_{p}$:
\begin{eqnarray}
\NB & = & P_2 - \min\left(P_2, \frac{p}{1-p}P_1 \right) \ECn_{p}
\end{eqnarray}
Hence, while directly related to a normalized EC, the NB looses one of its attractive qualities: that a value of 1.0 indicates a system that has the same performance as a naive system that does not use the input samples to make its decisions. For this reason, we believe normalized EC is preferable to NB.

\subsection{F-beta score}
\label{sec:fscore}
The F-beta score, which we will call $\FS$, is defined as 1 minus a metric called effectiveness, which was first introduced in \cite{vanRijsbergen79}. The $\FS$ is probably one of the most popular metrics for binary classification. It assumes that the problem is not symmetric and one of the classes is taken as the class of interest to be detected. Here, we will take the class of interest to be class $H_2$. $\FS$ is defined as follows:
\begin{eqnarray}
    \FS &=& (1+\beta^2) \frac{\Pre\Rec}{\beta^2\Pre+\Rec} 
\end{eqnarray}
where 
\begin{eqnarray}
    \Pre = \frac{N_{22}}{N_{*2}}, \ \ \ \
    \Rec = \frac{N_{22}}{N_{2*}} 
\end{eqnarray}
Replacing those values in $\FS$, we get:
\begin{eqnarray}
    \FS &=& \frac{ (1+\beta^2) N_{22}}{(1+\beta^2)N_{22} + \beta^2 N_{21} + N_{12}} \label{eq:fs}
\end{eqnarray}
$\FS$ takes values between 0 and 1. Larger values indicate better performance, contrary to the EC for which larger values indicate worse performance.
For this reason, in order to compare $\FS$ with EC, it is convenient to work with $1-\FS$: %. For convenience, we will name this metric $\OMFS$.
\begin{eqnarray}
    1-\FS &=& \frac{ \beta^2 N_{21} + N_{12}}{(1+\beta^2)N_{22} + \beta^2 N_{21} + N_{12}}\\
    &=& \frac{ \beta^2 N_{21} + N_{12}}{\beta^2 N_{2*} + N_{*2}}\\
    &=& \frac{ \beta^2 R_{21} P_2 + R_{12} P_1}{\beta^2 P_2 + R_{*2}} 
\end{eqnarray}
where $R_{*2} = N_{*2}/N$ is the fraction of samples in the evaluation set that are labelled as class $H_2$.
We can see that the numerator is the EC for the binary case when costs are given by $c_{11}=c_{22}=0$, $c_{12}=1$, $c_{21}=\beta^2$ and the priors are obtained from the evaluation data, which we will call $\ECb$:
\begin{eqnarray}
    \ECb &=& \beta^2 R_{21} P_2 + R_{12} P_1.  \label{eq:ecbeta}
\end{eqnarray}
Now, we can express $1-\FS$ as a function of $\ECb$ and its normalized version:
\begin{eqnarray}
    1-\FS &=& \frac{ \ECb}{\beta^2 P_2 + R_{*2}} \\
    & = & \min(\beta^2 P_2, P_1) \frac{ \NECb}{\beta^2 P_2 + R_{*2}}
    \label{eq:ec_vs_fs}
\end{eqnarray}
That is, $1-\FS$ is proportional to $\ECb$ with a scaling factor given by the inverse of $\beta^2 P_2 + R_{*2}$. The first term in that expression is independent of the system, it only depends on the prevalence of class $H_2$ in the evaluation dataset, $P_2$.
When comparing different systems on the same dataset (or on different datasets with the same class priors) $P_2$ is fixed.\footnote{Note that values of $\FS$ across datasets with different priors should not be compared with each other since the change in priors implies a change in the weight given to each type of error which is, essentially, a change in metric: comparing $\FS$ on two datasets with different priors is like comparing accuracy with balanced accuracy. Trying to infer which dataset is easier for the system based on that comparison would be unreasonable. The same, of course, goes for EC. Yet, for EC we can still do the comparison by using priors given by those we expect to see when the system is used rather than those observed in the evaluation datasets with the different priors. The EC computed with fixed costs and priors can then be safely compared across datasets \cite{godau2023deployment}. One could, perhaps, do a similar trick for $\FS$ but it is not worth the trouble since, as we will see next and in Section \ref{sec:ec_vs_fs}, $\FS$ has other problems that make it a poor metric.}
The second term in the denominator in \eref{eq:ec_vs_fs} is the percentage of samples which are labeled by the system as being of class $H_2$. Given two systems with the same value for $\ECb$, $\FS$ will favor the one that detects more samples as being from class $H_2$, regardless of whether they are correctly or incorrectly classified (examples of this behavior will be given in Section \ref{sec:ec_vs_fs}). This seems like an odd thing to reward, having more samples from class $H_2$ detected is not a good thing in itself. If we are more interested in detecting the samples from class $H_2$ than those of class $H_1$, the principled thing to do is use an EC which penalizes errors in samples from class $H_2$ more than errors in samples of class $H_1$ (i.e., setting $c_{21} > c_{12}$). 

As for the EC, we can compute the best $\FS$ that would correspond to a naive system that always outputs the same decision. If the system chooses always class $H_1$, $\FS=0$ since $N_{22}=0$. On the other hand, if the system chooses always class $H_2$, the recall is 1 and the precision is equal to the prior of class $H_2$, $P_2$, which results in $\FS=(1+\beta^2) P_2 / (\beta^2 P_2 + 1)$. This is the value of $\FS$ for the best naive system which always makes the same decision. 
Any system with an $\FS$ worse than this value should be considered ineffective.
We will not use a normalized version of the $\FS$ because this is not usual practice in the community, but we will use this value as a reference when looking at examples. 

\subsection{Matthews correlation coefficient}
\label{sec:mcc}

The Matthews correlation coefficient (MCC) was first introduced in \cite{matthews75} for comparison of chemical structures, later proposed as a metric for binary classification in \cite{Baldi00}, and finally generalized to the multi-class case in \cite{Gorodkin04}. In this section we will consider the binary version since this is enough for our purpose.
Chicco et.~al \cite{Chicco2020} claim that MCC is the most informative single score to establish the quality of a binary classifier that outputs hard decisions. EC is not considered in that paper as an option in the comparison between metrics. As we discuss below and in Section \ref{sec:ec_vs_mcc}, we believe EC is superior to MCC as a metric for classification performance.

The MCC for the binary case is defined as follows:
\begin{eqnarray}
    \MCC & =&  \frac{N_{11} N_{22} - N_{12} N_{21}} {\sqrt{(N_{11}+N_{21})(N_{11}+N_{12})(N_{22}+N_{12})(N_{22}+N_{21})}} \\
    & = & \frac{(N_{1*}-N_{12}) (N_{2*}-N_{21}) - N_{12} N_{21}} {\sqrt{N_{1*}N_{2*}N_{*1}N_{*2}}}\\
    & = & \sqrt{\frac{N_{1*}N_{2*}}{N_{*1}N_{*2}}}\  \frac{N_{1*}N_{2*}-N_{2*}N_{12} - N_{1*}N_{21}} {N_{1*}N_{2*}} \\
    & = & \sqrt{\frac{N_{1*}N_{2*}}{N_{*1}N_{*2}}}\  \left( {1-\frac{N_{12}}{N_{1*}} - \frac{N_{21}}{N_{2*}}} \right) \\
     & = & \sqrt{\frac{N_{1*}N_{2*}}{N_{*1}N_{*2}}}\  \left( {1- \left( R_{12} + R_{21}\right)} \right) 
\end{eqnarray}
Note that, unlike $\FS$ and EC, MCC can take negative values. Large positive values correspond to good systems with low error rates, while, as we will see next, negative values indicate systems that are worse than the naive one.

We can now see that the MCC is related to the $\ECn$ obtained by setting $\alpha=1$ in \eref{eq:ec2}, which we will call $\NECu$ (for ``uniform'' error weights).
Hence, we can now express MCC as a function of this EC:
\begin{eqnarray}
    \MCC & =& \sqrt{\frac{N_{1*}N_{2*}}{N_{*1}N_{*2}}}\  \left( {1- \NECu} \label{eq:ec_vs_mcc} \right) \label{eq:mcc}
\end{eqnarray}

From this equation we see that when the $\NECu$ is larger than 1, which, as explained in Section \ref{sec:norm_ec}, happens when the system is worse than a naive system that does not use the input samples, the MCC is negative. Further, MCC and $1-\NECu$ differ by the following factor:
\begin{eqnarray}
    \sqrt{\frac{N_{1*}N_{2*}}{N_{*1}N_{*2}}} =
    \sqrt{\frac{N_{1*}(N-N_{1*})}{N_{*1}(N-N_{*1})}} = \sqrt\frac{P_1 (1-P_1)}{R_{*1}(1-R_{*1})}
\end{eqnarray}
where $R_{*1}=N_{*1}/N$ is the fraction of samples in the evaluation set that are labeled as class $H_1$. 

The numerator in this factor is fixed, it does not depend on the system, only on the priors. Hence, it is irrelevant for comparing systems on the same dataset.
The denominator, on the other hand, does depend on the system and is largest when $R_{*1}=0.5$ and smaller as the decisions become more imbalanced (i.e., as $R_{*1}$ approaches 0 or 1). Hence, the factor will grow as the decisions become more imbalanced, making MCC larger in absolute value. For two systems with the same $\NECu<1$, MCC will favor the one with larger imbalance in the decisions. This does not seem like a reasonable behavior for a metric since the frequency with which a system makes a certain decision is not a good or bad quality in itself. Notably, even when $P_1=0.5$ this metric will penalize systems that have $R_{*1}$ close to 0.5. Therefore, as in the $\FS$ case, the MCC rewards arbitrary behaviors of the system that do not necessarily  indicate better performance. We will see examples of this undesirable behavior in Section~\ref{sec:ec_vs_mcc}.

\subsection{LR+}
The likelihood ratio for positive results, usually called LR+, is commonly used in diagnostic testing~\cite{Pauker75}. As for $\FS$, it is used for non-symmetric binary classification problems where one of the classes is the class of interest. It is given by:
\begin{eqnarray}
    \LR & = &\frac{\mathrm{sensitivity}}{1-\mathrm{specificity}} 
\end{eqnarray}   
where, if we assume class $H_2$ is the class of interest, $\mathrm{sensitivity} = R_{22}$ and $\mathrm{specificity} = R_{11}$.
LR+ is always positive and the larger its value, the better the system.

We can write LR+ as a function of an EC value using the fact that $R_{22}+R_{21}=1$, and $R_{11}+R_{12}=1$:
\begin{eqnarray}
 \LR & = & \frac{1-R_{21}}{R_{12}} \\
    & = & \frac{1-R_{21}-R_{12}}{R_{12}} + 1\\
    & = & \frac{1-\NECu}{R_{12}} + 1
\end{eqnarray}
So, given two systems with the same $\NECu$, the LR+ will favor the one with lower $R_{12}$. Note, also, that when the system is worse than the naive system in terms of $\NECu$, the LR+ is smaller than 1. For system that are better than the naive one, though, LR+ can take any value from 1 to infinity. We see no reason to prefer this metric over the EC where, if favoring lower values of $R_{12}$ is desired, one can simply set $c_{12}>c_{21}$.

\subsection{Examples}
\label{sec:examples_decisions}
In this section we show results for three different NECs, $\FS$ and MCC for a large set of different confusion matrices. We focus on $\FS$ and MCC because they are the ones for which the relationship with NEC is more complex and the examples can help better understand their behaviour.
We focus on three comparisons:
\begin{itemize}
    \item $\NECu$ versus MCC. These two metrics are related by \eref{eq:ec_vs_mcc}.
    \item $\NECb$ the normalized version of the EC in \eref{eq:ecbeta} versus $\FS$, with $\beta=1$ in both cases. These two metrics are related by \eref{eq:ec_vs_fs}. To make the value of $\beta$ explicit in the name, we call these metrics $\NECbO$ and $\FSO$. 
    \item $\NECb$ with $\beta=2$, which we call $\NECbT$, versus the $\FSO$. As we will see, in our examples, these two metrics correlate better than the two metrics above for systems that are better than the naive one.
\end{itemize}

Figure \ref{fig:ec_vs_other_scatter} shows those comparisons for two datasets: a balanced one with $P_1=0.5$ ($N_{1*}=N_{2*}=500$), and an imbalanced one with $P_1=0.9$ ($N_{1*}=900$, $N_{2*}=100$). In both cases $N_{12}$ and $N_{21}$ vary between 0 and their maximum possible value ($N_{1*}$ and $N_{2*}$, respectively) in 20 steps. For each combination of $N_{12}$ and $N_{21}$ we have a point in each of the plots. Tables \ref{tab:metric_comp_bal} and \ref{tab:metric_comp_imb} show the values of the metrics as well as the error rates and the fraction of samples detected as each class for a selection of these systems for the balanced and the imbalanced datasets, respectively. We discuss the  results in the plots and tables in the sections below.

\subsubsection{EC vs MCC}
\label{sec:ec_vs_mcc}

\begin{figure*}[t]
\centering
\includegraphics[width=\columnwidth]{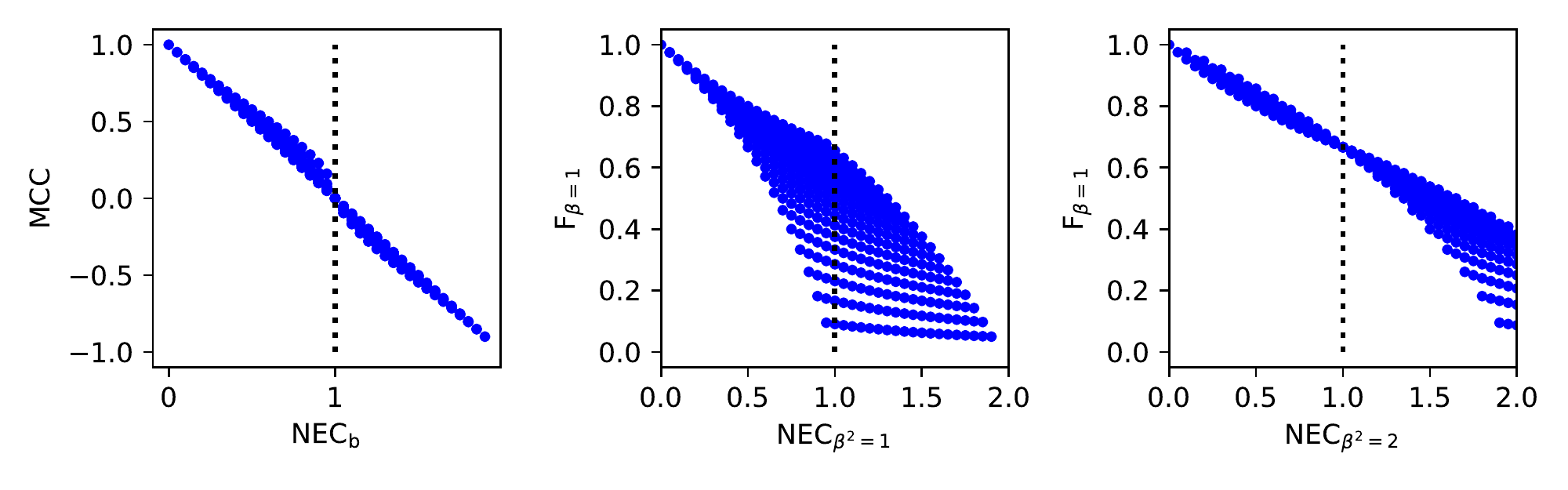}
\includegraphics[width=\columnwidth]{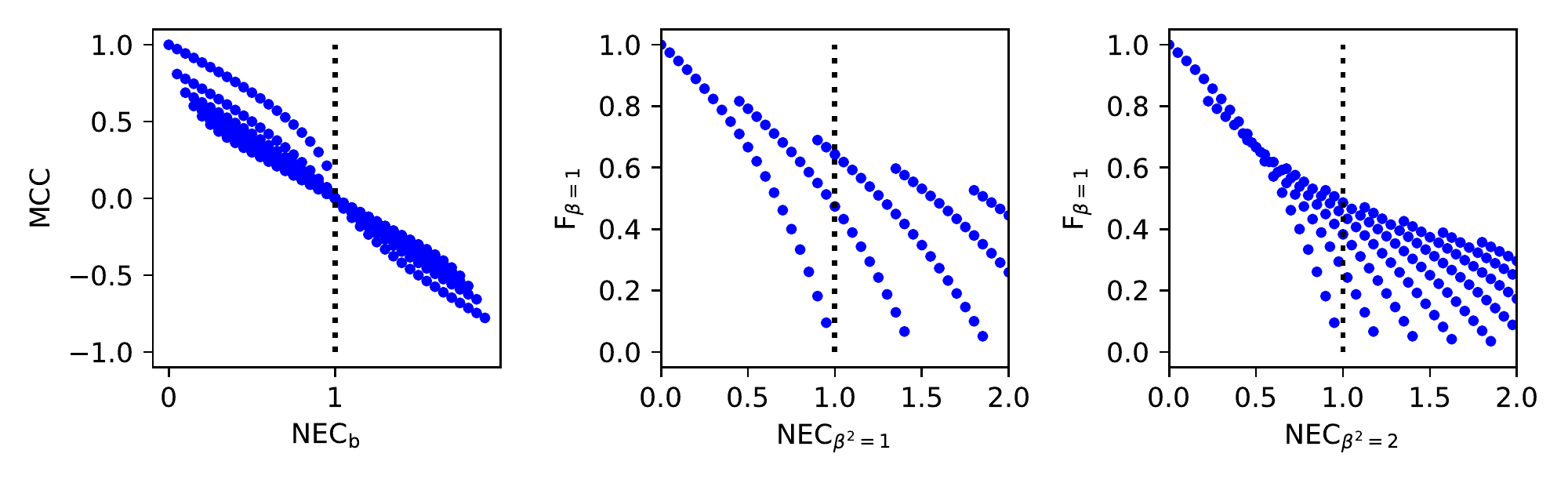}
\caption{Comparison of metrics for a balanced dataset ($P_1=0.5$) at the top, and the imbalanced dataset ($P_1=0.9$) at the bottom. The metrics are described in Section \ref{sec:examples_decisions}. The x-axis limits are restricted to a maximum of 2.0 to focus attention in that region of interest  (for the bottom $\FS$ plots, many values spill over to regions with larger NEC).}
\label{fig:ec_vs_other_scatter}
\end{figure*}

The left plots in Figure \ref{fig:ec_vs_other_scatter} show the comparison between $\NECu$ and MCC.
We can see that MCC and  $\NECu$ correlate reasonably well on these datasets, specially for the balanced one. Yet, detailed results in Table \ref{tab:metric_comp_bal} show that for the same $\NECu$ value the MCC tends to favor the more imbalanced systems, even on the balanced dataset (see the last block where the best MCC values correspond to the two imbalanced systems). Similarly, Table \ref{tab:metric_comp_imb} shows that MCC favors the more imbalanced systems even when the imbalance is opposite to the one in the dataset (see the case of $\NECu=0.90$ in the third block, where MCC is larger for the two most imbalanced systems). This is a clearly undesirable behaviour. In summary, as also discussed in Section \ref{sec:mcc}, the MCC behaves almost like $1-\NECu$ but with a preference toward systems with imbalanced decisions which makes it a poor metric compared to $\NECu$.

On a side note, staring at Table \ref{tab:metric_comp_bal} the reader might be concerned to see that systems with very imbalanced decisions may get the same $\NECu$ than systems with balanced decisions. For the balanced dataset, it seems like a good metric should be able to tell that a system with balanced decisions is preferable. The balanced and imbalanced systems are, in fact, quite different in nature. If we assume that the systems are making their decision for each sample based on a vector of scores (as in Figure~\ref{fig:architecture}), then it is likely that the system that  generates imbalanced decisions for the balanced dataset for the $\NECu$ metric is actually using a suboptimal decision-making procedure. This means that the $\NECu$ for such system could be easily improved by fixing the decision stage. On the other hand, the system that generates balanced decisions is probably already doing its best, given the scores. One of the advantages of using $\NECu$ as metric is that it enables this analysis. We will discuss these issues in depth in Section \ref{sec:metrics_on_scores}.

\begin{table}[ht]
%\footnotesize
    \centering
    \begin{tabular}{cc|ccccc|cc|cc}
$N_{21}$ & $N_{12}$ & $\NECu$ &  $\NECbO$ &  $\NECbT$ & $\FSO$ & MCC & $R_{21}$ & $R_{12}$ & $R_{*2}$ &  $R_{*1}$ \\
\hline
   0 &   50 &      0.10 &      0.10 &      0.10 &      0.95 &      0.90 &   0.00 &  0.10 &   0.55 &  0.45 \\
  25 &   25 &      0.10 &      0.10 &      0.15 &      0.95 &      0.90 &   0.05 &  0.05 &   0.50 &  0.50 \\
  50 &    0 &      0.10 &      0.10 &      0.20 &      0.95 &      0.90 &   0.10 &  0.00 &   0.45 &  0.55 \\
\hline
   0 &  250 &      0.50 &      0.50 &      0.50 &      0.80 &      0.58 &   0.00 &  0.50 &   0.75 &  0.25 \\
 125 &  125 &      0.50 &      0.50 &      0.75 &      0.75 &      0.50 &   0.25 &  0.25 &   0.50 &  0.50 \\
 250 &    0 &      0.50 &      0.50 &      1.00 &      0.67 &      0.58 &   0.50 &  0.00 &   0.25 &  0.75 \\
\hline
   0 &  450 &      0.90 &      0.90 &      0.90 &      0.69 &      0.23 &   0.00 &  0.90 &   0.95 &  0.05 \\
 225 &  225 &      0.90 &      0.90 &      1.35 &      0.55 &      0.10 &   0.45 &  0.45 &   0.50 &  0.50 \\
 450 &    0 &      0.90 &      0.90 &      1.80 &      0.18 &      0.23 &   0.90 &  0.00 &   0.05 &  0.95 \\

    \end{tabular}
    \caption{Comparison of metrics from different systems in the top row in Figure  \ref{fig:ec_vs_other_scatter} (i.e., on balanced data). The NEC values are better when lower, while MCC and $\FS$ are better when higher. Each block has a different fixed value of $\NECu$, which for this balanced dataset is equal to $\NECbO$.}
    \label{tab:metric_comp_bal}
\end{table}

\begin{table}[ht]
%\footnotesize
    \centering
    \begin{tabular}{cc|ccccc|cc|cc}
$N_{21}$ & $N_{12}$ & $\NECu$ &  $\NECbO$ &  $\NECbT$ & $\FSO$ & MCC & $R_{21}$ & $R_{12}$ & $R_{*2}$ &  $R_{*1}$ \\
 \hline   
   0 &   90 &      0.10 &      0.90 &      0.45 &      0.69 &      0.69 &   0.00 &  0.10 &   0.19 &  0.81 \\
   5 &   45 &      0.10 &      0.50 &      0.28 &      0.79 &      0.78 &   0.05 &  0.05 &   0.14 &  0.86 \\
  10 &    0 &      0.10 &      0.10 &      0.10 &      0.95 &      0.94 &   0.10 &  0.00 &   0.09 &  0.91 \\
\hline
   0 &  450 &      0.50 &      4.50 &      2.25 &      0.31 &      0.30 &   0.00 &  0.50 &   0.55 &  0.45 \\
  25 &  225 &      0.50 &      2.50 &      1.38 &      0.38 &      0.33 &   0.25 &  0.25 &   0.30 &  0.70 \\
  50 &    0 &      0.50 &      0.50 &      0.50 &      0.67 &      0.69 &   0.50 &  0.00 &   0.05 &  0.95 \\
\hline
   0 &  810 &      0.90 &      8.10 &      4.05 &      0.20 &      0.10 &   0.00 &  0.90 &   0.91 &  0.09 \\
  40 &  450 &      0.90 &      4.90 &      2.65 &      0.20 &      0.06 &   0.40 &  0.50 &   0.51 &  0.49 \\
  90 &    0 &      0.90 &      0.90 &      0.90 &      0.18 &      0.30 &   0.90 &  0.00 &   0.01 &  0.99 \\
\hline
   0 &   90 &      0.10 &      0.90 &      0.45 &      0.69 &      0.69 &   0.00 &  0.10 &   0.19 &  0.81 \\
  45 &   45 &      0.50 &      0.90 &      0.68 &      0.55 &      0.50 &   0.45 &  0.05 &   0.10 &  0.90 \\
  90 &    0 &      0.90 &      0.90 &      0.90 &      0.18 &      0.30 &   0.90 &  0.00 &   0.01 &  0.99 \\
    \end{tabular}
    \caption{Comparison of metrics from different systems in the bottom row in Figure \ref{fig:ec_vs_other_scatter} (i.e., on imbalanced data with $P_1=0.9$). The NEC values are better when lower, while MCC and $\FS$ are better when higher. The first three blocks have a fixed value of $\NECu$, while the bottom block has a fixed value of $\NECbO$}
    \label{tab:metric_comp_imb}
\end{table}

\subsubsection{EC vs $\FS$}
\label{sec:ec_vs_fs}

The second column of plots in Figure \ref{fig:ec_vs_other_scatter} show the comparison between the $\NECbO$ and $\FSO$. The scatter plot shows the effect of the denominator in \eref{eq:ec_vs_fs}. We can see that, for the better systems, with $\NECbO$ close to 0 and $\FSO$ close to 1, the two values are correlated. Yet, as the systems start to degrade, the metrics become less correlated. This means that if we are comparing two systems with each other, our conclusion of which system is best will strongly depend on the selected metric. 
Further, it is easy to find examples where the $\NECbO$ is close to the value for the best naive system for that metric, while the $\FSO$ is significantly better than the corresponding value. For example, when $P_1 = 0.1$, the best naive system for both metrics is the one that outputs always class $H_2$. Such a system has an $\NECbO = 1$ (Section \ref{sec:norm_ec}) and $\FSO = 0.18$  (Section \ref{sec:fscore}). Now, a system with $N_{21}=10$ and $N_{12}=90$ has a $\NECbO = 1$ and $\FSO = 0.62$. Hence, in terms of $\NECbO$ this system performs as badly as the system that always chooses class $H_2$, while in terms of $\FSO$ the system is much better than this naive system. This is because the $\FSO$ is rewarding this system for having a relatively large $R_{*2}$ of 0.18, compared to the prior for class $H_2$, which is 0.10. $\FSO$ is implicitely and indirectly favoring systems that prioritize not missing samples from class $H_2$. This can be done in a more direct and transparent way by choosing a NEC with a larger cost $c_{21}$.

To illustrate how a larger $c_{21}$ can achieve the desired behavior, the right-most column in Figure~\ref{fig:ec_vs_other_scatter} shows the comparison between $\FSO$ with $\NECb$ with $\beta^2=2$. We can see that, in the region corresponding to good systems, the correlation between the two metrics increases compared to the one obtained using $\NECb$ with $\beta=1$. 
In particular, the system mentioned above with $N_{21}=10$ and $N_{12}=90$ has a $\NECbT = 0.55$, a value significantly better than that of the best naive system, in agreement with $\FSO$. Yet, $\NECbT$ directly rewards this system for having a low error rate on samples of class $H_2$, rather than rewarding it for having a relatively large rate of samples classified as class $H_2$, as is the case for $\FSO$. While $\NECbT$ and $\FSO$ correlate relatively well, we strongly advise the reader to use $\NECbT$ (or some other NEC with a $\beta$ value designed for the problem of interest) rather than $\FSO$, which attempts to approximate the same behavior in an indirect way.

\subsection{Discussion on Metrics for Hard Decisions}

While the EC metric is explained in every statistical learning book, it unfortunately has not been widely adopted by the machine learning community. In this section we have argued that the NEC, the normalized version of the EC, is a better metric than the popular F-beta score, the MCC and other common classification metrics. We have also shown that it is a generalization of other two common metrics, the error rate and the balanced error rate which are, in turn, one minus the accuracy and balanced accuracy, respectively. The generalization is in the sense that costs and priors can be defined depending on the application of interest and that the set of decisions need not be the same as the set of classes. 

To summarize, the advantages of the EC over other metrics are as follows:

\begin{itemize}
    \item It has a straightforward intuitive interpretation. It is easy to see how the rate of each type of error and the priors affect its value.
    \item It lets developers assign costs for each type of error explicitly, which is useful when the errors are not all equally important. 
    \item It lets developers pick the priors independently from those observed in the data, which is useful when the priors in the evaluation data do not coincide with the ones expected in practice.
    \item It enables evaluation of systems for which the set of decisions do not coincide with the set of classes.
    \item It works out-of-the-box for the multi-class case.
    \item When normalized, the EC has a very intuitive range, with values above 1 indicating that the system is worse than a naive system that does not use the input samples to make its decisions.
    \item Finally, and quite importantly, it can be used to make optimal decisions and study calibration. These issues will be discussed in Section \ref{sec:bayesEC}.
\end{itemize}

Given all these advantages we argue that the EC, or its normalized version, NEC, should be the preferred metric for most (if not all) classification tasks when we want to evaluate hard decisions.

\section{Bayes Decision Theory}
\label{sec:bayes_decisions}

Before moving on to a discussion on metrics for scores, we need to review the basics of Bayes decision theory. Assume that we have selected a cost function for our problem of interest, $C(h,d)$, where $h$ is the true class of the sample and $d$ is a categorical or non-categorical decision. When $d$ is a categorical decision, the cost function reduces to the cost matrix we have been using in previous sections. When $d$ is not categorical, the cost function is a generalization of the cost matrix that allows for soft decisions. As we will see in Section \ref{sec:psrs}, this generalization of the cost matrix allows us to define metrics that evaluate the quality of the posteriors in an exhaustive way. 

In general, given the cost function, we will want to make optimal decisions in the sense that they minimize the expectation of the cost of interest, $\expv{C(h,d(x))}{}$, where we have explicitly added the dependency of $d$ on the input sample $x$. The expectation is taken with respect to a distribution over both $x$ and $h$, $p(h,x)$. Using the chain rule we can write $p(h,x) = p(x) P(h|x)$, which allows us to write the expectation of the cost as follows:
\begin{eqnarray}
\expv{C(h,d(x))}{p(h,x)} = \expv{\sum_{i=1}^K C(H_i,d(x)) P(H_i|x)}{p(x)} \label{eq:ec_gen}
\end{eqnarray}
The expression $\expv{v(z)}{p(z)}$ denotes the expected value of $v(z)$, where $z$ is one or more random variables, with respect to a distribution $p(z)$.
Note that while the expression above is an expected cost just like the EC from previous sections, in the rest of this paper we will still reserve the EC acronym for the special case in which decisions are categorical, as defined in Section \ref{sec:ec}.

The expectation in \eref{eq:ec_gen} can be minimized by choosing, for each $x$, the decision that minimizes the inner summation  \cite{bishop:ml,Hastie:statlearning}:
\begin{eqnarray}
 d_B(x) = \argmin_d \sum_{i=1}^{K} C(H_i,d) P(H_i|x) \label{eq:bayes_decisions}
\end{eqnarray}
The decision $d_B$ is usually called the \textit{Bayes decision}. If decisions are made in this way for every $x$, then the expected cost with respect to the distribution $p(h,x)$ will be minimized. 

Now, which distribution $p$ do we use to obtain this expectation and the Bayes decisions? We would like to use the ``true'' underlying distribution of our data. Unfortunately, unless we are running simulations, we do not and never will have access to such distribution. We can only obtain models for it. In fact, our classifier produces a model for $P(h|x)$. We can use those posteriors in \eref{eq:bayes_decisions} to make Bayes decisions. On the other hand, unless we do generative modeling, we usually do not have a model for the distribution of $x$. Fortunately, this distribution is not necessary for making Bayes decisions. 
It is needed, though, to compute the expected cost over the data, $\expv{C(h,d(x))}{p(h,x)}$. For the purpose of this work, when computing the expected cost we will  take $p(h,x)$ to be the empirical distribution in the evaluation data. That is, we will simply take the average cost over the evaluation data, as we did when defining the EC. 

\subsection{Calibration}
\label{sec:calibration}
Bayes decisions are guaranteed to minimize the expected cost when the expectation is taken with respect to the same posterior distribution used to determine those decisions, which in our case is given by the classifier's output scores. Note that this is true for any $p(x)$ since Bayes decisions do not depend on this distribution. Now, as mentioned above, in practice, after decisions are made using Bayes decision theory or any other decision-making approach, the expected cost is computed using the empirical distribution of the test data. The Bayes decisions may or may not be optimal for this empirical estimate of the expected cost. This leads us to the following definition for calibration: 
\begin{quote}
{\bf A classifier is calibrated (for the selected cost) if the Bayes decisions made with its outputs are optimal in the sense that they minimize the average cost on the evaluation data. That is, if there is no better way to make decisions using the classifier's output than Bayes decision theory.}
\end{quote}
More formally, we say that a model that outputs scores $s(x) = P(.|x) = \{P(H_i|x) \mid i=1,\ldots,K\}$ is calibrated for the cost function $C$ if the following condition is satisfied:
\begin{eqnarray}
    \expv{C(h,d_B(s(x)))}{} \leq \expv{C(h,d_B(\tilde s(x)))}{} \label{eq:calib_def_bayes}
\end{eqnarray}
where $\tilde s(x) = f(s(x))$, and $f$ is a transformation of the vector of posteriors output by our model into a new vector of posteriors. Hence, if no transformation of the scores results in a lower expected cost than the original scores, we say that scores are calibrated. In this work, we assume the expectations on both sides are taken with respect to the empirical distribution, though other models of the distribution could be used for this purpose.

One $f$ that minimizes the expected cost with respect to the empirical distribution is the one that maps the score $s$ for each $x$ in the evaluation dataset to the one-hot vector with a one at the true class for that $x$ and zeros elsewhere. This procedure assumes there are no two samples with the same $x$ value but different class labels, which is a reasonable assumption for most practical problems when $x$ is continuous and multi-dimensional. The resulting mapping would lead to a cost of zero. Yet, of course, when $x$ is continuous, such mapping is unlikely to generalize well to any other dataset (in fact, $s(x)$ is not even defined for any $x$ that is not in the evaluation dataset). Hence, such a mapping is not helpful. Our goal should be to find an $f$ that improves the expected cost on a yet unseen dataset, the data the system will encounter when deployed. To this end, $f$ should be learned on held-out data to avoid overfitting the test data which could lead us to wrongly conclude that the system is miscalibrated.

This definition of calibration is not the one commonly used in the literature. Instead, the usual definition of calibration is to say that a system is perfectly calibrated if the score vector it produces satisfies the following equality:
\begin{eqnarray}
    s_i = P(H_i|s)\ \  \forall i \in \{1, \ldots, K\}. \label{eq:calib_def}
\end{eqnarray}
that is, if the score for every class $H_i$ coincides with the posterior probability of class $H_i$ given the score vector generated by the system. The intuitive interpretation of this definition is that for a well calibrated system, we would expect that, among all the samples for which the system outputs a score for class $H_i$ of $r$, $r$ of them would be of class $H_i$. When this definition is used, $P(h|s)$ is referred to as the ``true'' or ``underlying'' distribution of the classes given the scores. Yet, as we discussed for $P(h|x)$, this distribution is not known except in simulations. In practice, when this definition is used, a model for $P(h|s)$, equivalent to the $\tilde s$ above, is created and a distance between this model and $s$ is computed. We will explain the most widely used approach of this family of metrics, the expected calibration error (ECE), in Section \ref{sec:ece}.

Importantly, both definitions of calibration are conceptually equivalent and declare that a system is miscalibrated if it is possible to find a better set of posteriors by transforming the original posteriors output by the system. Yet, the first definition above explicitly states what ``better'' means: that the expected cost for the transformed scores is lower than that of the original scores. Further, it makes explicit the fact that our assessment of calibration will necessarily depend on our best efforts in finding the transformed or calibrated posteriors as well as on the cost function of interest.

\subsection{Special cases}
\label{sec:bayes_decisions_special}
A special scenario of Bayes decision theory arises when $\mathcal D = \mathcal H$ and the cost matrix is the 0-1 square matrix. That is, when $C(h,d)=0$ if $h=d$, and $C(h,d)=1$ if $h\neq d$. In this case the EC reduces to one minus the accuracy (see Section \ref{sec:accuracy}) and the Bayes decisions reduce to
\begin{eqnarray}
 d_B(x) & = & \argmin_h \sum_{i=1|H_i\neq h}^{K} P(H_i|x)\\
 & = & \argmin_h 1- P(h|x)\\
 & = & \argmax_h P(h|x).
 \label{eq:map_decisions}
\end{eqnarray}
which is obtained by replacing the 0-1 cost function defined above in \eref{eq:bayes_decisions}.
This is the standard argmax decision rule used in most of the literature in machine learning for making decisions in a multi-class setting. We can now see that this rule is only optimal for the 0-1 cost matrix. In any other scenario, if the costs are not equal to 1 for all error types or if the cost matrix is not square (i.e., if the set of decisions is not the same as the set of classes), then \eref{eq:bayes_decisions} should be used to make decisions using the corresponding cost matrix.

Another special case of \eref{eq:bayes_decisions} corresponds to binary classification when $\mathcal H = \mathcal D$ and the cost matrix is square with zeros in the diagonal. In that case, the equation reduces to:
\begin{eqnarray}
 d_B(x) = \begin{cases}
    H_2, & \text{if } c_{12} P(H_1|x) < c_{21} P(H_2|x) \\
    H_1,              & \text{otherwise}\label{eq:bayes_decisions_bin0}
\end{cases}
\end{eqnarray}
Using the fact that $P(H_2|x)=1-P(H_1|x)$, this rule becomes:
\begin{eqnarray}
 d_B(x) = \begin{cases}
    H_2, & \text{if } P(H_2|x) > c_{12}/(c_{12}+c_{21})  \\
    H_1,              & \text{otherwise}
\end{cases}  \label{eq:bayes_decisions_bin}
\end{eqnarray}
That is, for binary classification with one decision per class, Bayes decisions simply consist of comparing the posterior for one of the classes with a specific threshold which is a function of the costs. While to optimize metrics like the $\FS$ or MCC we need to sweep a threshold and choose the best one for that metric, for the EC we simply need to plug the costs into \eref{eq:bayes_decisions_bin} to get the optimal threshold. Note that this is true, as long as the classifier is calibrated for the selected cost, in which case Bayes decisions are indeed optimal. If this is not the case, the threshold above would be suboptimal and it would have to be tuned. Alternatively, the posteriors can be transformed or \emph{calibrated} to enable the use of Bayes decision theory, as will be described in Section \ref{sec:calibration_loss}.

\subsection{Posterior Dependence on Class Priors}

Bayes rule states that:
\begin{eqnarray}
P(h|x)  =  \frac{p(x|h)P(h)}{p(x)} =  \frac{p(x|h)P(h)}{\sum_j p(x|H_j) P(H_j)}. \label{eq:post_from_lks}
\end{eqnarray}
That is, the posterior probability for class $h$ depends on the likelihoods $p(x|h)$ and the prior probabilities $P(h)$ for all classes.  
As a consequence, the posteriors generated by a classifier depend on the class priors corresponding to the data used for training the system unless a class-weighted objective function was used for training, in which case the priors are determined by those weights. When the test priors, i.e., those in the evaluation data or those selected to compute the EC, are different from those used in the training process, the decisions made with \eref{eq:bayes_decisions} may be suboptimal because the posteriors are calculated with a mismatched prior distribution. In other words, the posteriors generated by the classifier may be miscalibrated.

For these cases in which the training and test priors are potentially mismatched, we would ideally like to have access to the likelihoods $p(x|h)$ with which we can compute the posteriors for the test priors, $P(h)$, using Bayes rule above\footnote{This, of course, assumes that the likelihoods remain valid across the two datasets even when the priors change, which may or may not be the case depending on the scenario.}. 
Now, the problem is that many modern machine learning approaches output posteriors, not likelihoods. In this case, a simple workaround is to calibrate the posteriors generated by the system using a function trained on data with the relevant priors (or, alternatively, setting the objective function to simulate these priors). The calibration process will be explained in Section \ref{sec:calibration_loss}.

Alternatively, for the special case of binary classification, the posteriors can be obtained as:
\begin{eqnarray}
P(H_2|x) & = & \frac{1}{1+\frac{P(H_1)}{P(H_2)}\frac{1}{\mathrm{LR}(x)} }, \label{eq:post_from_llrs}\\
P(H_1|x)& = & 1-P(H_2|x), \nonumber
\end{eqnarray}
where 
\begin{eqnarray}
\mathrm{LR}(x) = \frac{p(x|H_2)}{p(x|H_1)}. \label{eq:lr}
\end{eqnarray}
That is, for this case, all we need to obtain posteriors given a certain set of target priors is the ratio between the two likelihoods. This approach is commonly used in speaker verification where the target priors are usually not known when training the classifier. In this task, the classifier is generally trained to directly output LRs by integrating the computation of the posterior, \eref{eq:post_from_llrs}, in the computation of the objective function. This can be done directly when training the classifier, or in a separate calibration stage.
For more on this approach, see, for example \cite{brummer2013likelihood}.

Conveniently, in the particular case of binary classification with $\mathcal H = \mathcal D$ and a square cost matrix with zeros in the diagonal, we do not need to go through the step of finding the posteriors in order to apply Bayes decision theory. The decision rule in \eref{eq:bayes_decisions_bin0} can be converted to use the LR directly as follows:
\begin{eqnarray}
 d_B(x) = \begin{cases}
    H_2, & \text{if } \mathrm{LR}(x) > \frac{c_{12} P(H_1)}{c_{21} P(H_2)} \\
    H_1,              & \text{otherwise}\label{eq:bayes_decisions_llrs}
\end{cases}
\end{eqnarray}

In summary, the quality of Bayes decisions will directly depend on the quality of the posteriors. These posteriors should, among other things, be computed using priors matched to the target test priors. Hence, when the system outputs posteriors, we need to make sure that the priors implicit in those posteriors coincide with the test priors. If this is not the case, we should take steps to solve this mismatch by either having the model output likelihoods or likelihood ratios (for the binary case) and using them to compute posteriors for the target priors, or by transforming (calibrating) the posteriors output by the system to a new set of posteriors adapted to those priors. Section \ref{sec:calibration_loss} discusses diagnosis and correction of calibration issues, including those caused by a mismatch in priors.

\section{Metrics for System Scores}
\label{sec:metrics_on_scores}

In Section \ref{sec:metrics_on_decisions} we discussed metrics that evaluate performance of hard decisions. Assuming the decisions are made based on a set of scores, as in Figure \ref{fig:architecture}, one might, instead, wish to evaluate the performance of the scores directly, independently of the decision-making stage. In this section, we review different metrics that can be used for assessing performance of scores. 

\subsection{Expected Value of Proper Scoring Rules}
\label{sec:psrs}

In this section we will describe a family of metrics which are expected values of proper scoring rules (PSRs) \cite{properscoring,brummer_thesis}. PSRs measure the goodness of the scores assuming they are posterior probabilities and that they will be used for making Bayes decisions. Given a score vector $s$ output by the system for a sample $x$ of class $h$, and a cost function $C(h,d)$ which represents the cost incurred by the system when making decision $d$ for a sample of class $h$, we can construct a PSR, $C^*$, as follows:
\begin{eqnarray}
 C^*(h,s) = C(h,d_B(s)) \label{eq:psr_const}
\end{eqnarray}
where $d_B(s)$ is the Bayes decision corresponding to cost function $C$. Note that we have expressed $d_B$ as a function of $s$ rather than $x$ as in Section \ref{sec:bayes_decisions}, since we have shown that Bayes decisions only depend on the vector of posteriors $P(.|x)$ which here we take to be equal to $s(x)$, the classifier's output.
The PSR, $C^*$, measures the goodness of the posteriors by putting them to work, using them to make Bayes decisions for the given cost function, which are the best possible decisions if all we know are the posteriors returned by the system. 

The PSR measures the cost for a single sample. To obtain a metric we can use for evaluation, we need to compute the expectation of the PSR over the data. As we did for the EC, in practice, we generally use the empirical estimate of the expectation. We will use EPSR to refer to the empirical estimate of the expectation of a PSR:
\begin{eqnarray}
 \EPSR = \frac{1}{N} \sum_{t=1}^N C^*(h^{(t)},s^{(t)}).
\end{eqnarray}
As we will see below, different $\EPSR$s can be obtained by using different cost functions $C$. 

\eref{eq:psr_const} is a constructive definition of PSRs. The usual definition, though, is that a function of $h$ and $s$ is a PSR if it satisfies that its expected value with respect to a distribution $q(h)$ over the classes is minimized if the score vector $s$ coincides with this distribution \cite{DawidMusio2014}. That is, $C^*$ is a PSR if it satisfies
\begin{eqnarray}
    q \in \argmin_s \expv{C^*(h,s)}{q(h)}. \label{eq:psr_property}
\end{eqnarray}
This property is satisfied for any $C^*$ constructed with \eref{eq:psr_const}, since the Bayes decisions for $s$ are those that minimize the expected cost with respect to $s$. Hence, the expectation of a PSR constructed with \eref{eq:psr_const} is minimized when $s=q$.
Theoretical details and a thorough analysis of PSRs can be found in, for example, in \cite{properscoring,brummer_thesis}. 

Note that \eref{eq:psr_property} is not an equality. The expected value of a PSRs may be minimized not only for the distribution $q$ but also for other distributions. When only $q$ minimizes the expected value, the PSR is called strict. Below, we introduce one non-strict PSR, the Bayes EC, and two strict PSRs, cross-entropy and Brier score.

\subsubsection{Bayes EC}
\label{sec:bayesEC}

When we take the decision $d$ to be categorical, the EPSR corresponding to the resulting cost function is simply the EC from \eref{eq:ec} when the decisions are made using Bayes decision theory, i.e., using \eref{eq:bayes_decisions}. We will call this EPSR, Bayes EC or $\ECB$.\footnote{This metric is sometimes called Bayes risk in some text books \cite{duda}. Here, we choose to continue using the term EC to highlight the fact that it is simply the EC for Bayes decisions.} The Bayes EC evaluates the quality of the scores assuming they will be used for making hard decisions using Bayes decision theory for a single operating point. The operating point is defined by the selected cost matrix and priors which are, in turn, used to determine the region corresponding to each decision. 
One special case of this metric is the accuracy computed using argmax decisions (see Section \ref{sec:bayes_decisions_special}), perhaps the most common metric used in the literature for multi-class classification.

Note that $\ECB$ is a non-strict PSR. It does not evaluate the quality of the scores across their full range of values (the simplex in $\mathbb{R}^K$) since its value only depends on the Bayes hard decisions. If we have two sets of posteriors which happen to result in the same Bayes decisions, they will have the same $\ECB$, but one of them may be better than the other for a different set of costs or priors.  Hence, $\ECB$ is only appropriate for measuring the quality of the scores if the application of interest is well defined and we know it will not change in the future. On the other hand, if the application of interest is not well determined, if there is more than one possible application, if the priors may vary across datasets or samples, or if we want to have interpretable scores as the system's output, we need to resort to strict PSRs which are only minimized at the reference distribution  \cite{brummer_thesis}. The following two sections give two examples of strict PSRs.

\subsubsection{Cross-entropy}
Taking $d$ to be a \emph{soft} decision given by a vector of posteriors that lives in the simplex in $\mathbb{R}^K$, we can define a cost function $C(h,d) = -\log(d_h)$, called the logarithmic loss, where, in a slight abuse of notation, $d_h$ is the vector $d$ at the index corresponding to $h$ (i.e.,  if $h=H_i$ $d_h = d_i$). 
It can be shown \cite{brummer_thesis} that, for this cost function, the Bayes decision is $d_B(s) = s$. That is, the (soft) decision that minimizes the expected logarithmic-loss is the input score itself. Hence, the EPSR resulting from this cost function is also the logarithmic loss. 
The expectation of this loss with respect to the empirical distribution of the evaluation data results in the cross-entropy, which is widely used in machine learning as objective function for system training:
\begin{eqnarray}
\XE  =  -\frac{1}{N} \sum_{t=1}^{N} \log(s_{h^{(t)}}(x^{(t)})) \label{eq:crossent_ave}
\end{eqnarray}
where, with the same notation as above, $s_{h^{(t)}}(x^{(t)})$ is the system's output for sample $x^{(t)}$ for class $h^{(t)}$, the true class of sample $t$.
This expression can be rewritten in a similar way as we did for the $\EC$ in \eref{eq:ec}:
\begin{eqnarray}
\XE  & = & - \sum_{t=1}^{N} \frac{P_{h^{(t)}}}{N_{h^{(t)}}} \log(s_{h^{(t)}}(x^{(t)})) \\
& = & -\sum_{i=1}^{K} \frac{P_i}{N_i} \sum_{t=1|h^{(t)}=H_i}^{N} \log(s_i(x^{(t)})). \label{eq:crossent}
\end{eqnarray}
In this expression, as in the $\EC$ case, we can see that the prior $P_i = P(H_i)$ could potentially be set to a value different from the prior in the evaluation data. As was discussed in the context of the $\EC$, this is useful when the priors in the evaluation data are not the ones we expect to see when the system is deployed. 

If $s_{h^{(t)}}(x^{(t)})$ is close to 1 for every $t$, then XE will be low. If, on the other hand, $s_{h^{(t)}}(x^{(t)})$ is close to 0, even for a single sample, the XE can grow to very large values. That is, this metric penalizes heavily systems that are very certain about an incorrect decision. This is a desirable property in many applications. For this reason, the $\XE$ is an excellent default EPSR for evaluating the goodness of systems that output posterior probabilities.

The logarithmic loss is a strict PSR, as can be shown directly by using the fact that the Kullback-Liebler discrepancy between two distribution is only zero when the distributions are equal \cite{DawidMusio2014}. 
A less well-known fact is that the XE can be derived as an integral over a family of $\ECB$ values where the costs are parameterized in a specific way \cite{brummer_thesis}. That is, the cross-entropy is, in fact, a generalization of the $\ECB$ in the sense that, instead of considering a single operating point, the $\ECB$ for all possible operating points are integrated with non-zero weights. 
The fact that weights are non-zero over the whole simplex makes the resulting PSR strict, as shown in \cite{brummer_thesis}.
By using XE as an evaluation metric we are assessing the ability of the system for making Bayes decision across the whole range of possible operating points rather than committing to a single one as when using the $\ECB$. 

As for the EC, we can obtain a normalized version of the XE by dividing its value with the XE of the best naive system (i.e., one that ignores the input $x$). The best naive system for any EPSR is given by the one that always outputs the prior distribution (see Section 2.4.1 in \cite{brummer_thesis}). The XE of this system is given by $-\sum_{i=1}^{K} P_i \log(P_i)$, which is the entropy of the prior distribution. Dividing the XE by this entropy gives us an interpretable metric for which a value larger than 1.0 indicates that the system is worse than a naive system.

\subsubsection{Brier Score}
\label{sec:brier}

Setting again $d$ to be a vector in the simplex in $\mathbb{R}^K$ but this time selecting the cost function to be 
\begin{eqnarray}
C(h,d) = \frac{1}{K} \sum_{i=1}^K (d_i-I(h=H_i))^2
\end{eqnarray}
we get another PSR. Here, $I(h=H_i)$ is the indicator function as defined under \eref{eq:conf_matrix}. As for the logarithmic-loss, it can be shown that the Bayes decision for this loss is the score itself. Hence, again, the PSR coincides with the cost function. 
Averaging this cost over the data we get the Brier score:
\begin{eqnarray}
\BR  =  \frac{1}{N} \sum_{t=1}^{N} \frac{1}{K} \sum_{i=1}^K (s_i(x^{(t)})-I(h^{(t)}=H_i))^2 \label{eq:brier_ave}
\end{eqnarray}
For each sample, this loss measures the average square distance to 1 for the posterior for the true class of the sample and to 0 for all other classes. 

As for the EC and the XE, we can express BR as an explicit function of the priors.
\begin{eqnarray}
\BR  & = & \sum_{t=1}^{N} \frac{P_{h^{(t)}}}{N_{h^{(t)}}} \frac{1}{K} \sum_{i=1}^K (s_i(x^{(t)})-I(h^{(t)}=H_i))^2  \label{eq:brier}
\end{eqnarray}
which allows us to redefine the priors to the test values if they do not coincide with the ones in the evaluation data.
For the binary classification case, since $s_1(x)=1-s_2(x)$ and $I(h=H_1)=1-I(h=H_2)$, the  score simplifies to:
\begin{eqnarray}
\BR & = & \sum_{t=1}^{N} \frac{P_{h^{(t)}}}{N_{h^{(t)}}} (s_2(x^{(t)})-I(h^{(t)}=H_2))^2.\label{eq:brier_bin}
\end{eqnarray}

As for XE, it can be shown that BR is a strict PSR \cite{DawidMusio2014} and that it can be obtained as an integral over a family of $\ECB$ \cite{brummer_thesis} with non-zero weight over the full simplex. What varies between both cases is the weight given to each point in the simplex. As a consequence of the difference in weighting, XE and BR penalize non-perfect posteriors differently.
Notably, the main difference is that BR is bounded. The maximum penalization it gives is 1, which happens when the  posterior for the true class is 0. On the other hand, as discussed above, XE gives infinite penalization to those errors, making it a better metric for many applications where extreme errors could have extreme consequences. For example, consider a medical application where the task is to detect the presence of a malign tumour from a medical image. A system that, even if rarely, may output a probability of tumour of 0.0 for a patient that does have a tumour should probably not be used in practice.

Finally, as for the XE, a normalized version of the Brier score can be computed by dividing by the Brier score of a naive system that always outputs the priors, $P_i$. We can compute the Brier score for this system by setting $s_i = P_i = N_i/N$ for every $x^{(t)}$. After some  manipulations, we get that the Brier score for this naive system is given by $1/K \sum_{i=1}^{K} P_i (1-P_i)$. 

\subsection{Calibration Loss using the Expected Value of PSRs}
\label{sec:calibration_loss}

As we saw in Section \ref{sec:bayes_decisions}, Bayes decision theory tells us how to make decisions that minimize the expected cost with respect to a certain distribution. As discussed in Section \ref{sec:calibration}, in practice, Bayes decisions may not optimize the expected cost of interest on the evaluation data in which case we say that the classifier is misscalibrated. 
This definition of calibration directly suggests a simple, flexible and general-purpose metric to evaluate calibration called calibration loss. This metric has been used in speech processing tasks that involve binary classification like speaker verification or language detection for almost two decades (see, for example, \cite{brummer:2006}). Here we describe a general version of the calibration loss used for those tasks which can be applied to multi-class classification \cite{Kull2015}.

{\bf A miscalibrated classifier is one for which the posteriors cannot be trusted to make optimal decisions using Bayes decision theory}. For a miscalibrated classifier, it is possible to find some way to make decisions based on the classifier's output that leads to better expected cost than if Bayes decisions were used. Equivalently, we can say that a system is miscalibrated if it is possible to transform the classifier's output so that the Bayes decisions taken with those new outputs lead to better expected cost than the when using the original outputs. Hence, a simple way to assess a system's miscalibration is to compute the difference between the expected cost when using the original posteriors to make Bayes decisions and when using transformed posteriors, where the transformation is trained to minimize the EPSR for the cost of interest. This transformation of the posteriors aimed at improving the EPSR is called calibration\footnote{Hence, the term calibration may refer both to a property of a classifier and to a transform applied to its output.}.

Specifically, calibration loss is obtained by the following simple procedure:
\begin{itemize}
    \item First, put the original scores to work. Measure their performance using one of the EPSRs explained in the sections above (or any other EPSR). Call this metric EPSRraw.
    \item Second, transform the scores using some calibration procedure.
    \item Third, put the calibrated scores to work and measure again their performance using the same EPSR. Call this metric EPSRemin, where emin stands for \emph{estimated minimum}, since this is our best estimate of the minimum value of EPSR that can be obtain by transforming the scores.
    \item Finally, compute the calibration loss, given by the difference between EPSRraw and EPSRemin:
    \begin{eqnarray}
    \mathrm{CalLoss} = \mathrm{EPSRraw}-\mathrm{EPSRemin}
    \label{eq:calloss}
    \end{eqnarray}
    Alternatively, a relative version of this metric can be computed:
    \begin{eqnarray}
    \mathrm{RelCalLoss} = 100\ \frac{\mathrm{EPSRraw}-\mathrm{EPSRemin}}{\mathrm{EPSRraw}} \label{eq:relcalloss}
    \end{eqnarray}
\end{itemize}
Using \eref{eq:calloss} we can now express EPSRraw as the sum of two terms, EPSRemin and CalLoss. EPSRemin is the part of EPRSraw that was not reduced by doing calibration. This term represents the discrimination performance of the system measuring how well the system separates the classes. The CalLoss term, on the other hand is the part of EPRSraw that is due to misscalibration. This decomposition is extremely useful during development as it allows us to diagnose where the attention should be focused if we want to improve performance of the system: a system with large calibration loss can be improved easily by adding a calibration stage after the classifier. In fact, the same calibration transform used to compute the calibration loss can be used. On the other hand, if the calibration loss is low, but the EPSRemin term is large, the development effort should be focused on improving the classifier itself since a simple transformation of the scores will not help improve the EPRS.

Note that this decomposition depends on the choice of calibration approach. A better approach will lead to a larger estimate of the CalLoss, while a bad one could even result in a negative value of the CalLoss. While having the metric depend on the choice of calibrator might be unsatisfying, it is an issue that is inherent to the problem of calibration assessment. Any metric designed to measure calibration quality needs to compute a model for the posterior distribution $P(h|s)$ in some way, implicitly or explicitly, choosing assumptions about the form of the posterior and a dataset on which to estimate its parameters (the dataset is usually chosen to be the test data itself, though as we will discuss below this standard choice is far from ideal). Calibration loss makes these choices completely flexible and explicit, which we believe to be one of the advantages of this approach. 

Readers familiar with calibration may be wondering how the calibration loss computed as in \eref{eq:calloss} relates to the calibration-refinement decomposition proposed by DeGroot in \cite{DeGroot83}. For calibration loss, the  EPSR of the raw scores is decomposed in two terms where the discrimination component is given by the EPSR of the calibrated scores. DeGroot, on the other hand, decomposes the EPSR of the raw scores with respect to the distribution defined by the calibrated posteriors, with the discrimination component (or refinement, in his terminology) given by the EPSR of the calibrated posteriors with respect to those same posteriors (see Theorem 4 in \cite{DeGroot83}). Hence, both approaches assume a calibration transformation is available, but they use it differently to obtain the total loss and its decomposition. In practice, in our experience, when the calibration transformation works well, the two decompositions lead to similar or identical results. This analysis is out of the scope of this paper and will be the focus of a future publication. 

\subsubsection{Calibration Approaches}

A detailed discussion of calibration methods is out of the scope of this paper, but we will mention some of the most standard approaches and provide references for further reading. Perhaps the most standard calibration approach is to apply an affine transformation to the logarithm of the posterior vector and train the parameters by minimizing the cross-entropy which, as we have seen, is an EPSR. Instances of this approach are linear logistic regression, also known as Platt scaling for binary classification \cite{plattscaling}, an extension of Platt scaling for the multi-class case \cite{brummer:2006}, and temperature scaling which is a single-parameter version of Platt scaling \cite{guo:17}. 
In this work, we will use the affine transformation proposed in \cite{brummer:2006} where the transformed scores $\tilde s$ are given by
\begin{eqnarray}
    \tilde s = \mathrm{softmax}(\alpha \log(s) + \beta) \label{eq:affcal}
\end{eqnarray}
where $s$, $\tilde s$, and $\beta$ are vectors of dimension $K$, the number of classes, and $\alpha$ is a scalar. When taking $\beta=0$, this transformation reduces to temperature scaling. 
Generative approaches have also been proposed for calibration. In these cases, the distribution of scores for each class is computed using the data under some set of assumptions, and these distributions are then used to compute calibrated scores \cite{brummer2014generative,cumani2019tied}. 

Note that calibration should be treated as just another stage inside the classifier block in Figure~\ref{fig:architecture}, with its own parameters and hyperparameters that need to be trained and tuned, respectively. As such, care needs to be taken that the data used to train the calibration model is not used for evaluation of the performance of the resulting scores, i.e., to compute EPSRemin. Further, it is usually a good idea to train the calibration stage using data that was not used to train the classifier, since the scores on that data are likely biased, specially when the model overfitted the training data. Hence, the usual approach is to held-out some data when training the classifier to use as training data for the calibration stage. 

We would like to point out that much of the literature in calibration actually uses ``train-on-test'' calibration, where the calibration model is trained on the evaluation data when computing the calibration metric. This is problematic, even when calibration is only done for metric computation, since this transformation can, as any other part of the model, overfit to the training data. Hence, evaluating on the same data where calibration was trained may lead us to conclude that the gap between the raw and the calibrated scores is larger than it actually is, an issue that has been recently discussed in \cite{gruber2022} with regards to the standard ECE metric.
In binary classification, when the calibration transformation is highly restricted, the train-on-test strategy could be acceptable when enough data is available. This is, in fact, the standard procedure used when computing calibration loss for speaker verification. Yet, we do not recommend this approach in general and particularly not when the number of classes is large or the number of evaluation samples is small.

A special case of the calibration loss above is obtained for a binary classification problem when using the $\ECB$ as EPSR, with a square cost matrix with zeros in the diagonal. In this case, as we saw in Section \ref{sec:bayes_decisions}, Bayes decisions are simply obtained by comparing the posterior for class $H_2$ with a threshold given by the costs. In this particular case, instead of calibrating the scores, we can calibrate the threshold. That is,  we can
sweep the threshold and choose the one that gives the best EC. This threshold selection should be done on held-out data to avoid optimistic results. Given EPSRraw obtained by applying the threshold from \eref{eq:bayes_decisions_bin} and EPSRemin obtained with the threshold selected by optimizing the EC of interest, we can compute CalLoss. This value will tell us how calibrated the scores are for the particular operating point determined by the EC. This simple procedure, though, is only applicable for the EC for binary classification with the square cost matrix with zeros in the diagonal. For every other case, the CalLoss needs to be obtained through calibration of the posteriors.

\subsection{Expected Calibration Error}
\label{sec:ece}
The expected calibration error (ECE) \cite{naeini2015obtaining} was proposed as metric to assess the calibration quality of binary classification systems. To compute the ECE, the posteriors for class $H_2$ for all the samples in the evaluation dataset are binned into $M$ bins. 
The ECE is then computed as:
\begin{eqnarray}
\ECE = \sum_{m=1}^M \frac{|B_m|}{N}\left| \mathrm{class2}(B_m) - \mathrm{avep2}(B_m) \right| \label{eq:ecebin}
\end{eqnarray}
where $B_m$ is the set of samples for which the posterior for class $H_2$ is in the $m$th bin, $|B_m|$ is the number of samples in that bin, $\mathrm{class2}(B_m)$ is the fraction of samples of class $H_2$ within that bin, and $\mathrm{avep2}(B_m)$ is the average posterior for class $H_2$ for the samples in that bin. 

This definition was later modified so that it could be applied to multi-class classification \cite{guo:17}. In this case, the metric evaluates the confidences rather than the posteriors. The confidence for a certain sample $t$ is the posterior corresponding to the class selected by the system which, in the literature that uses ECE is taken to be the one with the maximum posterior. Hence, the confidence is simply the maximum posterior for a sample. This effectively maps the multi-class problem into a binary classification problem: deciding whether the system was correct in its decision (this is now called class $H_2$) by using its confidence as score. Given this new problem, we can now use the definition for the ECE, where $\mathrm{class2}$ is now the fraction of samples correctly classified by the system and $\mathrm{avep2}$ is the average confidence. This is usually written as:
\begin{eqnarray}
\ECEmc = \sum_{m=1}^M \frac{|B_m|}{N}\left| \mathrm{acc}(B_m) - \mathrm{conf}(B_m) \right|
\end{eqnarray}
where we have renamed the ECE as ECEmc to differentiate it from the original definition for binary calibration. 

\subsubsection{Calibration Loss using Brier loss versus ECE}

In section \ref{sec:calibration_loss} we defined CalLoss as the part of the EPSR that cannot be reduced by calibration. Here, we will compare this metric for calibration quality with the ECE metric defined above. We will do it for the binary case, for which the ECE can be computed without mapping the problem to an auxiliary binary problem. The CalLoss based on Brier score for binary classification is given by:
\begin{eqnarray}
 \mathrm{CalLoss}_\mathrm{BR} & = & \frac{1}{N} \sum_{t=1}^{N} \left[ (s_2(x^{(t)})-I(h^{(t)}=H_2))^2 -  (\tilde s_2(x^{(t)})-I(h^{(t)}=H_2))^2 \right] \label{eq:abscallossbrier}
\end{eqnarray}
where $s_2(x^{(t)})$ and $\tilde s_2(x^{(t)})$ are the raw and calibrated posteriors for class $H_2$ for sample $t$. Note that $\tilde s^{(t)}$ is a function of $s^{(t)}$ given by a calibration transformation. The expression shows that the absolute calibration loss is the average over all samples of a function of the scores before and after calibration. For samples of class $H_1$, the function is simply given by the difference between the square of the raw and calibrated posteriors of class $H_2$, $s_2(x^{(t)})^2 -  \tilde s_2(x^{(t)})^2$. For samples of class $H_2$, the distance is given by the difference between the square of the raw and calibrated posteriors for class $H_1$, $s_1(x^{(t)})^2 -  \tilde s_1(x^{(t)})^2$, since in that case $s_2(x^{(t)})-I(h^{(t)}=H_2) = s_2(x^{(t)})-1 = s_1(x^{(t)})$, and  $\tilde s_2(x^{(t)})-I(h^{(t)}=H_2) = \tilde s_1(x^{(t)})$.

The ECE for the binary case in \eref{eq:ecebin} can be rewritten as:
\begin{eqnarray}
\ECE & = & \frac{1}{N} \sum_{m=1}^M \ \sum_{t|s_2(x^{(t)})\in B_m} \left| \mathrm{class2}(B_m) - \mathrm{avep2}(B_m) \right| 
\end{eqnarray}
where we replaced the $|B_m|$ with a sum over all samples in that bin. We can now define $\hat s_2(x^{(t)}) = \mathrm{class2}(B_{m_t})$, where $B_{m_t}$ is the bin to which sample $t$ belongs, which is determined based on $s_2(x^{(t)})$. 
This function can be seen as a calibration transformation: it takes the input score $s_2(x^{(t)})$  and maps it to a new score given by the fraction of samples of class $H_2$ that have raw scores within the same bin. This is, in fact, a calibration method called histogram binning \cite{histogramBinning,naeini2014binary}. 
Using this definition, we can write the ECE as:
\begin{eqnarray}
\ECE & = &  \frac{1}{N} \sum_{t=1}^N \left| \hat s_2(x^{(t)}) - b(s_2(x^{(t)}))  \right| \label{eq:ece_bh_sample}
\end{eqnarray}
where $b(s_2(x^{(t)}))=\mathrm{avep2}(B_{m_t})$ is the average posterior over the bin corresponding to sample $t$, a quantized version of $s_2(x^{(t)})$.

Comparing \eref{eq:ece_bh_sample} with \eref{eq:abscallossbrier} we can see that while the form of these functions is different, they are conceptually very similar: both are averages of some function of the raw and the calibration scores over all samples in the test data.

\subsubsection{Discussion on ECE's drawbacks}
\label{sec:ece_criticism}

The comparison above may seem to suggest that ECE is as good as CalLoss for computing the quality of calibration. In our opinion, this is not the case. In this section we will present several theoretical reasons why we believe ECE is an inferior metric compared to CalLoss offering no advantage over it other than its popularity. In the experimental section we will further show empirical evidence of ECE's flaws.

In our view, the goal of computing a measure that purely reflects calibration quality is to make sure we are doing the best possible job given the scores provided by the system. As explained in Section \ref{sec:calibration_loss}, if the CalLoss is large, it means we can improve the  EPSR quite easily by adding a calibration transform as a final stage. On the other hand, if we find that CalLoss is low but the EPSRraw is high, we may conclude that the scores are simply not very discriminative. In this case, the problem needs to be solved by working on the classifier itself, usually a much more demanding endeavor than adding a calibration stage to the system. The ECE does not facilitate this analysis since it does not directly indicate how much would the system's overall performance improve by improving calibration. Is an ECE of 10 a problem? How much better would our posteriors be if we added a calibration stage to our system? This question cannot be responded by looking at ECE, while it is explicitly addressed by the relative CalLoss metric which is the percentage of the EPSRraw that can be reduced by doing calibration.

Another problem with ECE is that the calibration transformation used to compute it, which as we saw is histogram binning, is trained on the test data itself. If this transformation overfits the data, either because the test dataset is too small or because we have chosen to use too many bins, the ECE will be overestimated, as discussed in \cite{gruber2022}. The fact that the transformation is trained on the test data also means there is no mechanism for tunning the number of bins since ECE will, in most cases, get smaller and smaller as we reduce the number of bins. In the extreme, if we use a single bin and the average posterior matches the prior in the test data (which usually happens if the priors in the training data are equal to those in the test data), the ECE would be 0. For this reason, in the literature, the value of $M$ is simply set arbitrarily to some low value, usually between 10 and 15. Yet, those numbers could be too small for some datasets leading to an underestimation of the calibration loss, and they could still be too big for very small datasets. Alternative versions for the ECE have been proposed where the bins are adapted to contain equal number of samples, showing that the resulting ECE is more robust \cite{nixon2019measuring}. Yet, this only mitigates the problem, without fully solving it. A simple solution to this problem is to estimate the calibration transformation on data other than the test data. If the transformation overfitted or underfitted, it will be bad on the test data, leading to a small CalLoss estimate. The usual system development effort can then be applied to the calibration approach, searching for better transforms until the developer is satisfied to have done the best effort. The resulting CalLoss can then be taken as a measure of how much the system would gain from calibration which, in turn, tells us how miscalibrated it is. We would like to note, though, that in most (if not all) scenarios the authors have encountered, the simple affine calibration approach described in Section \ref{sec:calibration_loss} does an excellent job at calibrating the system. Additional efforts in searching for better calibration approaches usually lead to marginal gains. Hence, affine calibration often suffices to diagnose and fix calibration problems in a system. 

A third issue with ECE is that, as explained above, while ECE for binary classification measures the quality of the posteriors, the ECEmc for multi-class classification as defined above measures only the quality of the confidences, ignoring all other values in the posterior vector. It assumes that the decisions will be made by selecting the class with the maximum score. Yet, as we have seen in Section~\ref{sec:bayes_decisions}, these decisions are only optimal when all costs are equal. Hence, if we aim to make optimal decisions for cost matrices that are not the 0-1 matrix, or if we want interpretable posteriors, then we should evaluate the full score vector and not just its maximum value. In the examples section we will show a case in which ECE fails to detect some severe case of misscalibration due to this issue.
In some works, including \cite{guo:17}, the multi-class definition of ECE is used for binary problems, which is quite unfortunate since the original definition for binary classification does not suffer from the problem of evaluating only confidences. Finally, a recent work has proposed a generalization of the ECEmc that evaluates the full posterior \cite{widmann2019}. Unfortunately, this metric is not based on EPSRs and does not provide a decomposition of the overall performance, suffering from the same issues mentioned above for the ECE. 

Another, perhaps more minor, issue with ECE is the fact that it does not penalize extreme errors (cases in which the true class has probability 0) severely enough, something we have also discussed with regards to the Brier score. As for Brier score, for ECE, a sample for which the posterior for the true class is 0 can only contribute as much as 1 to the average. 

Note that, if histogram binning is the desired calibration approach, perhaps for its conceptual simplicity, our recommendation is to train this transformation on held out data and then compute calibration loss using an  EPSR, with cross-entropy being a good general choice. In addition, if overfitting is of no concern, perhaps because the dataset is very large, then the transformation could be trained on the test data itself. This specific implementation of calibration loss using histogram binning trained on the test data has exactly the same assumptions as the ECE but has the advantage of providing a decompositions of the overall EPSR, turning it into a more interpretable and actionable metric. 

In summary, we believe the ECE, while easy to compute and intuitive, has too many disadvantages and no advantages (other than tradition) with respect to the more elegant and well-motivated calibration loss based on PSRs. We will show a comparison of results for these metrics in Section \ref{sec:examples_scores}.

\subsection{Metrics for Assessing Interpretability of Scores}

Much of the literature on calibration centers around the issue of interpretability. In most works, interpretable scores are taken to be those that are well-calibrated. This view of interpretability implicitly assumes that given two systems A and B, if A has a lower calibration error then it is preferable (more interpretable) than B. 
The problem with this approach is that calibration error reflects only one part of the performance of the posteriors. Consider an extreme example where system A is a naive classifier that outputs the class priors for every sample. Such a system would have a calibration error of 0 but would not be useful for interpretation as it has no information about the input samples. 

When we are looking for interpretable scores we should aim for the best possible posteriors, $P(h|x)$, those that are closest to the distribution of our data. This is exactly what EPSRs measure. Hence, given the two systems A and B above, if A has a lower calibration error but a larger EPSR than B, we should say that system B is more interpretable because its posteriors are better. In addition, knowing that B is misscalibrated we should add a calibration stage to the system which would give us an even better EPSR. We see, then, that there is no reason to choose a system with lower calibration error over one with lower EPSR. 

If not useful for assessing interpretability, then what is the purpose of calibration metrics? We believe that the goal of calibration analysis should only be to guide the development process. Should we or should we not bother to include a calibration stage in our system. This can be answered by computing a calibration-only metric. Other than for this diagnosis, calibration-only metrics play, in our view, no role in the evaluation of a system. If good posteriors are required, then EPSRs should be used for evaluation.

\subsection{Area Under the Curve and Equal Error Rate}

For binary classification problems, evaluation of the scores is greatly simplified by the fact that it suffices to evaluate one of the two scores, since they both sum to one. Assuming that decisions will be made by thresholding this score, we can sweep the threshold across the range of possible values and create two types of curves:
\begin{itemize}
    \item Receiver Operating Characteristic (ROC) curves, which correspond to the $R_{22}$ rate versus the $R_{12}$ rate 
    \item Recall vs Precision (PR) curve 
\end{itemize}
For each of these two curves, the area under the curve (AUC) can be obtained as a summary metric. Further, from the ROC we can find the threshold for which $R_{12}$ equals $R_{21}=1-R_{22}$. The value of $R_{12}$ at this threshold is called Equal Error Rate (EER). Interestingly, there is a very close tie between the EER and the $\ECB$ computed with the 0-1 cost matrix: this $\ECB$ is upper bounded by $\min(\mathrm{EER}, P_1, P_2)$ \cite{brummer21_interspeech}, as long as calibration is perfect. If calibration is not perfect, $\ECB$ can grow without bound, while EER will not.

These ROC and PR curves and the resulting metrics (AUCs and EER) are immune to any monotonic transformation of the scores. Since most calibration approaches for binary classification assume monotonicity (it usually does not seem reasonable or desirable to change the ranking of samples with the calibration process), calibration does not affect these metrics. In other words, these metrics ignore the issue of threshold selection which implicitly means they assume that, if hard decisions need to be made, the selection can always be done optimally. This may or may not be a reasonable assumption in practice, depending on the task, the application, and the available data. Further, note that, since these metrics are immune to calibration problems, they do not measure whether the scores can be used to make good Bayes decisions or whether they are interpretable as posterior probabilities. 
It is important to recognize these limitations: these metrics only tell one part of the story on the performance of the system. If one wishes to assess the effect of threshold selection, use the system for making Bayes decisions, or obtain interpretable scores, AUC and EER are not sufficient for performance evaluation.

Finally, note that these metrics cannot be naturally generalized to the multi-class case, or to binary cases in which the set of decisions do not coincide with the set of classes. In those cases, we believe EPSRs are the best option for performance evaluation. 

\subsection{Specificity at Fixed Sensitivity}

An approach used for measuring performance in some binary classification applications \cite{Bickelhaupt18} is to fix the sensitivity (or recall, or, in our terminology, $R_{22}$) to some pre-defined value and report the specificity ($R_{11}$) corresponding to the threshold that results in that value of sensitivity.  This metric, which we will call $\SPEC$, is used when the application of interest imposes a certain target sensitivity below which the system would be unacceptable. It compares systems by forcing them to operate at that exact level of sensitivity and comparing their specificity at that threshold.
Sometimes, the opposite is done, fixing the specificity or one minus specificity ($R_{12}$) and reporting sensitivity \cite{biamby2022}. The analysis below generalizes trivially to such metric.

$\SPEC$, while intuitive and easy to explain, has a number of problems. First, consider the following scenario. We have a development dataset where we compute $\SPEC$ for a fixed sensitivity of 0.95, which corresponds to a certain threshold $T_\mathrm{dev}$. Now, when this threshold is used on a new dataset, the sensitivity will most likely no longer be exactly our target value. Now, what is the value of our metric for this new dataset? Do we measure the specificity at the $T_\mathrm{dev}$ threshold? That would be wrong since the sensitivity at that threshold is not our target value. Do we change the threshold to the one for which the sensitivity is 0.95 in the new dataset? This is the usual way in which this metric is implemented: a new threshold is estimated for every dataset or system for which the metric is calculated. Yet, this approach is also problematic because, in practice, the threshold cannot be selected on the same data on which the system is being used, it needs to be predefined and applied blindly during deployment. A metric that is aimed at reflecting the performance that the system will have in practice should take this issue into account.

Note also that by computing the threshold to achieve the desired sensitivity on each new dataset or system we are implicitly considering different costs for each new case. Given a certain system and dataset, the threshold used to compute $\SPEC$ corresponds to the best threshold for a certain EC. That is, when setting the threshold we are, implicitly, optimizing a certain EC with its corresponding costs for the two types of errors. Now, if we change the system or the evaluation dataset, the threshold to achieve the target sensitivity will change and it will correspond to a different EC, with different costs. This means that, when using the $\SPEC$ metric we are, in fact, changing our implicit assumption about the cost for each type of error. This does not seem to be a reasonable approach for any application. 

Another issue with this metric is that it forces all systems to operate at the minimum acceptable sensitivity. When comparing various systems with each other, better systems could potentially operate at better sensitivity values and still achieve good specificity. Yet, this metric does not allow us to adapt the threshold to obtain a better trade off between the two types of error when possible.

In some applications, though, we do need to impose a minimum value of sensitivity. The approach we recommend for these cases is to use a NEC as defined in \eref{eq:ec2}, where the $\alpha$ is determined using some baseline system to achieve the desired target sensitivity. The procedure would be as follows. Given a certain development dataset and baseline system:
\begin{itemize}
    \item Swipe $\alpha$ across a range of values
    \item For each value of $\alpha$, evaluate the corresponding NEC (\eref{eq:ec2}) for the development dataset for all possible decision thresholds and choose the threshold that minimizes that NEC. By choosing the decision threshold in this way, we make this step independent of possible calibration issues. 
    \item For each $\alpha$ we then have an optimal threshold and can measure the sensitivity achieved at that threshold on the development dataset.
    \item We can now choose the $\alpha$ that leads to the target sensitivity. We use NEC with that value of $\alpha$ as our metric.
\end{itemize}
This procedure translates the original requirement on sensitivity to a NEC that we can now use to compare systems with each other, select thresholds, and evaluate performance on different datasets without all the problems described above related to $\SPEC$. 
We note, though, that it is possible that for some system or dataset other than the baseline system and development dataset used to obtain $\alpha$, optimizing the selected NEC may result in a sensitivity lower than the target one. In that case, one can either go back to choosing the value of $\alpha$ for the new system or dataset and adopting the new resulting NEC as metric, or simply declare that system to be inadequate for the task.

As an added benefit, this approach makes explicit the implicit assumption on  the costs that come with the selected target sensitivity. To find what the ratio of costs corresponding to the selected $\alpha$ is, we simply observe that $c_{12}/c_{21} = \alpha P_2/P_1$, where the priors are those corresponding to the development set. We can then analyze whether the ratio that resulted from setting the sensitivity to the selected target value is, indeed, reasonable for the task. Say, for example, that we are setting the sensitivity to 0.95 because we believe it is very important to never miss a sample of class $H_2$. Now, after doing the procedure above we find that, for our baseline system and our development dataset, the ratio of costs for which that sensitivity is achieved is $c_{12}/c_{21} = 2$. That is, for the desired operating point with sensitivity equals to 0.95 we are, implicitly, assigning more weight to the errors on class $H_1$ than to those of class $H_2$. Upon seeing this, we might decide the selected target sensitivity was, in fact, not a good choice.

\subsection{Average over Class Detectors}

Some of metrics described above are only defined for binary classification. One way to be able to use these metrics for multi-class classification is to cast the original problem into one of multiple one-vs-other detection problems. That is, for each class $H_i$, we consider the system's output for that class as the score for a binary classifier where the two classes are  ``the sample is of class $H_i$'' and ``the sample is of any other class''. For each of these binary problems, a binary classification metric can be computed. The final metric is computed as the average of all these metrics. 

The problem with this approach is that it implicitly changes the task of interest, which we assume is classification, to one of multiple detections, one for each class. This task allows for zero, one or more detections to occur. That is, a sample could be labelled with no classes or with more than one class. This is something that would not be allowed in the original classification problem. Hence, any metric computed this way may or may not correlate with the actual performance the system will have once used to make classification decisions. We have no guarantees that the metric is in any way useful for our purposes. For this reason, we do not recommend the use of this type of average binary metrics when the actual task of interest is classification. Of course, there are many scenarios where the task of interest is, in fact, one of making potentially multiple binary detections. In this cases using averages of binary metrics would be appropriate.

\subsection{Examples}
\label{sec:examples_scores}

In this section we will show a comparison of metrics computed directly over scores, as well as, for the binary case, curves of metrics for hard decisions obtained by sweeping a threshold over the scores. For these experiments we use two simulated test sets for 2 and 10 classes, obtained with the following procedure:

\begin{enumerate}
    \item Set the class priors to $P_1=0.9$ and $P_i=0.1/(K-1)$ for $2\leq i\leq K$.
    \item Set the total number of samples, $N$, to 100000, and determine the number of samples for each class, $N_i$ as the closest integer to $P_i N$.
    \item Generate $N_i$ samples using a unidimensional Gaussian distribution with mean $\mu_i = i-1$ and standard deviation $\sigma_i =0.15$. We take this to be $x$, a unidimensional input feature from which we need to obtain the posteriors.
    \item Rather than training a system to obtain the posteriors or likelihoods, we directly use the generating distributions to compute the likelihoods $p(x|H_i) \sim \mathcal{N}(\mu_i, \sigma_i^2)$. These likelihoods are perfectly calibrated, meaning that, if used to obtain posteriors with the correct priors using \eref{eq:post_from_lks} they would lead to perfectly calibrated posteriors since they are obtained from the true underlying distribution of each class.
    \item Finally, assume two possible prior distributions: 1) the same distribution with which the data was generated, and 2) a mismatched distribution where the last class has a prior of 0.9 and the rest have prior $0.1/(K-1)$. 
    \item Using those two prior distributions, compute two sets of posteriors which we call Datap-cal and Mismp-cal which correspond to using the data priors and the mismatched priors to obtain the posteriors using \eref{eq:post_from_lks}. The cal in the posterior names refer to the fact that calibrated likelihoods are used to obtain them.
    \item For the binary case, we also compute the likelihood ratios using \eref{eq:lr}. We call these values LR-cal.
\end{enumerate}

Note that with the procedure above, Datap-cal is perfectly calibrated for the test data. Mism-cal is not, because, even though the likelihoods used to compute it are perfectly calibrated, the priors are mismatched to the ones used for testing. Further, to show the effect of other types of misscalibration, we create a misscalibrated version of the likelihoods by scaling and shifting the log-likelihoods obtained in the fourth step above. The scale is set to 0.5 for all classes and the shift is 0.5 for the first class and 0 for the others. Then, we repeat the procedure in steps 6 and 7 using the misscalibrated likelihoods to get Datap-mc1, Mism-mc1, and LR-mc1. Finally, we create a misscalibrated version of the posteriors by scaling by 0.2 the posteriors Datap-cal and Mism-cal in the log domain. After scaling the log posteriors, we renormalize them to obtain a second set of misscalibrated posteriors Datap-mc2, Mism-mc2.

In summary, after this process we have six different sets of posteriors (Datap/Mismp-cal/mc1/mc2) and two sets of LRs (cal/mc1) which we use to compute the various metrics in the experiments below.

\subsubsection{Threshold Selection for Binary Classification}
For binary classification the usual approach for making hard decisions is the following: (1) select a metric of interest, (2) sweep a threshold over the possible range of values for the score, computing this metric for each threshold on some development dataset, (3) select the threshold that results on the best value for the selected metric and apply it to the evaluation data. In many papers, the development set is taken to be the evaluation set itself. This is a  poor evaluation procedure when the goal is to estimate the future performance of the system, since it can lead to optimistic results. For our purposes in this section, though, for simplicity we will show performance on the same set where the threshold is estimated since the goal is to analyze the impact of the approach used for threshold selection rather than predicting future performance. 

For this section we use the LR-cal and LR-mc1 scores and show results for the same metrics used for the examples in Section \ref{sec:examples_decisions}:
\begin{itemize}
    \item $\NECu$ defined in Section \ref{sec:mcc},
    \item $\NECb$, the normalized version of \eref{eq:ecbeta} with $\beta=1$, $\NECbO$
    \item $\NECb$ with $\beta=2$, $\NECbT$, and
    \item $\FS$ from \eref{eq:fs} with $\beta=1$.
\end{itemize}
The posteriors used to compute the NECs are obtained with \eref{eq:post_from_llrs} using the target priors which are uniform for $\NECu$ and the ones in the data (0.9 for class $H_1$ and 0.1 for class $H_2$) for $\NECbO$ and $\NECbT$. 

The left plot in Figure \ref{fig:metrics_vs^{(t)}hr} shows the values of these metrics for LR-mc1 with hard decisions obtained for a range of different decision thresholds. We can see that, in this case, the thresholds determined by Bayes decision theory, \eref{eq:bayes_decisions_llrs} (dashed vertical lines) lead to suboptimal values of the corresponding NEC metric. For example, while the best $\NECbT$ is 0.366, the $\NECbT$ for the Bayes threshold is 0.604. This indicates the scores are poorly calibrated since for well-calibrated scores the Bayes threshold would coincide with the optimal one. A similar, though not as extreme, effect can be seen for $\NECbO$ and $\NECu$. The best threshold for $\FSO$ is close to the best threshold for $\NECbT$, in agreement with our observation in Section \ref{sec:examples_decisions} where we saw  that these two metrics are highly correlated when the system has good performance. Note, though, that by using $\NECbT$ we can do the calibration analysis, comparing the performance on the Bayes threshold with the performance on the best threshold. This cannot be done using $\FSO$.

If we only care about one specific NEC corresponding to a certain cost matrix and priors, then all we need to do in this case in which the scores are poorly calibrated is ignore Bayes decision theory and simply pick the threshold that leads to the best NEC of interest. This approach is completely valid, as long as we do not need interpretable scores at the output, and as long as we are willing to retune the threshold if we ever need to change the cost matrix or the priors for a new use-case scenario.

A more elegant and general solution to the problem we see in the left plot in Figure \ref{fig:metrics_vs^{(t)}hr} where the Bayes thresholds are not optimal, is to calibrate the scores. In a practical scenario we would have to train a calibration transformation to obtain these calibrated scores (Section \ref{sec:examples_cal} shows examples where calibration transformations are trained).
In our simulated example, though, we have the calibrated scores available by construction.
The right plot in Figure \ref{fig:metrics_vs^{(t)}hr} shows the different metrics computed on the calibrated scores, LR-cal. We can see that, in this case, the Bayes and the best threshold for each NEC fall very close together, resulting in almost identical NEC for both thresholds.  This indicates that, at least for the operating points corresponding to these three NECs, the scores are well calibrated. Further, as for the raw scores, the best $\NECbT$ and $\FSO$ result in a very similar threshold selection.

\begin{figure*}[t]
\centering
\includegraphics[width=0.49\columnwidth]{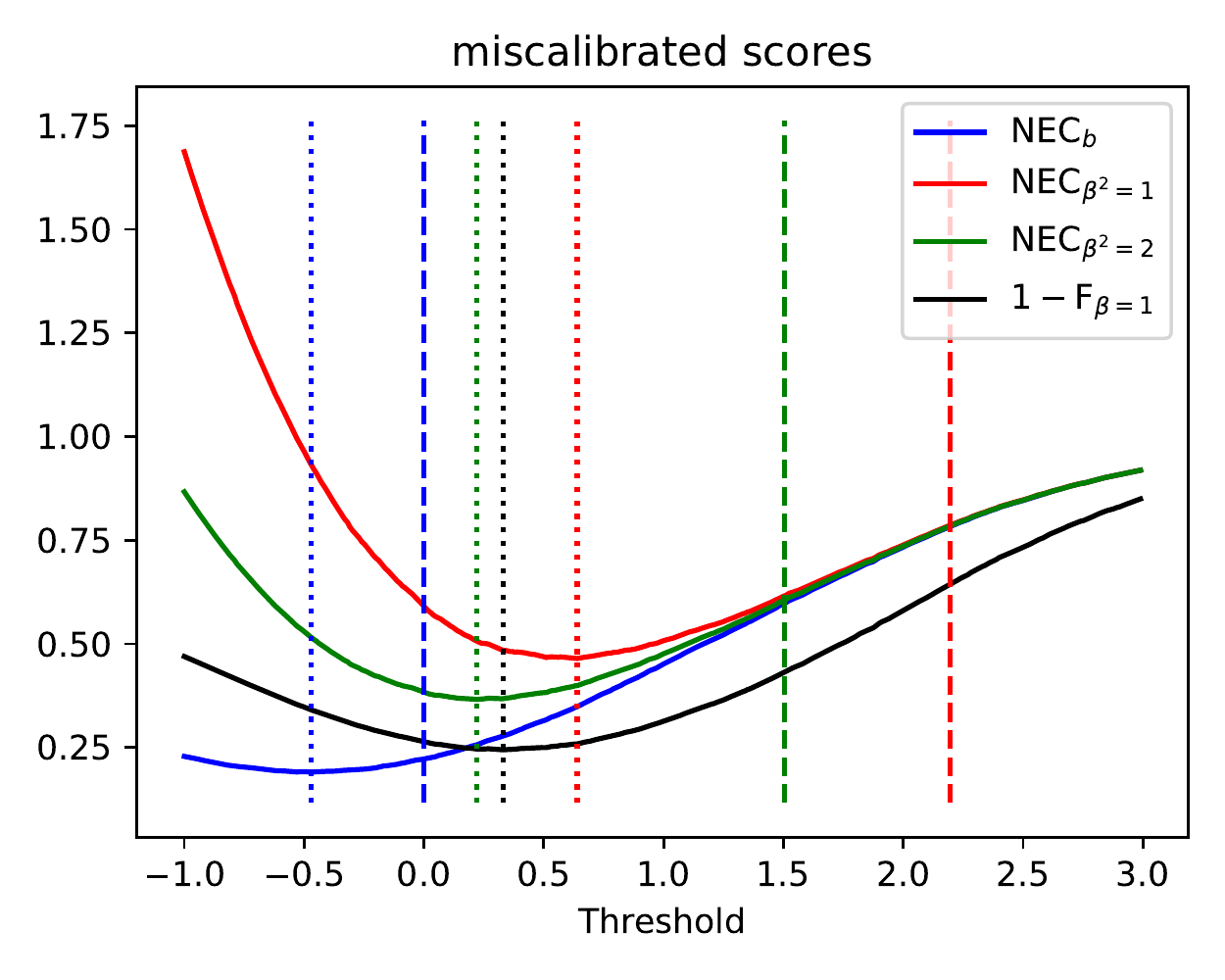}
\includegraphics[width=0.49\columnwidth]{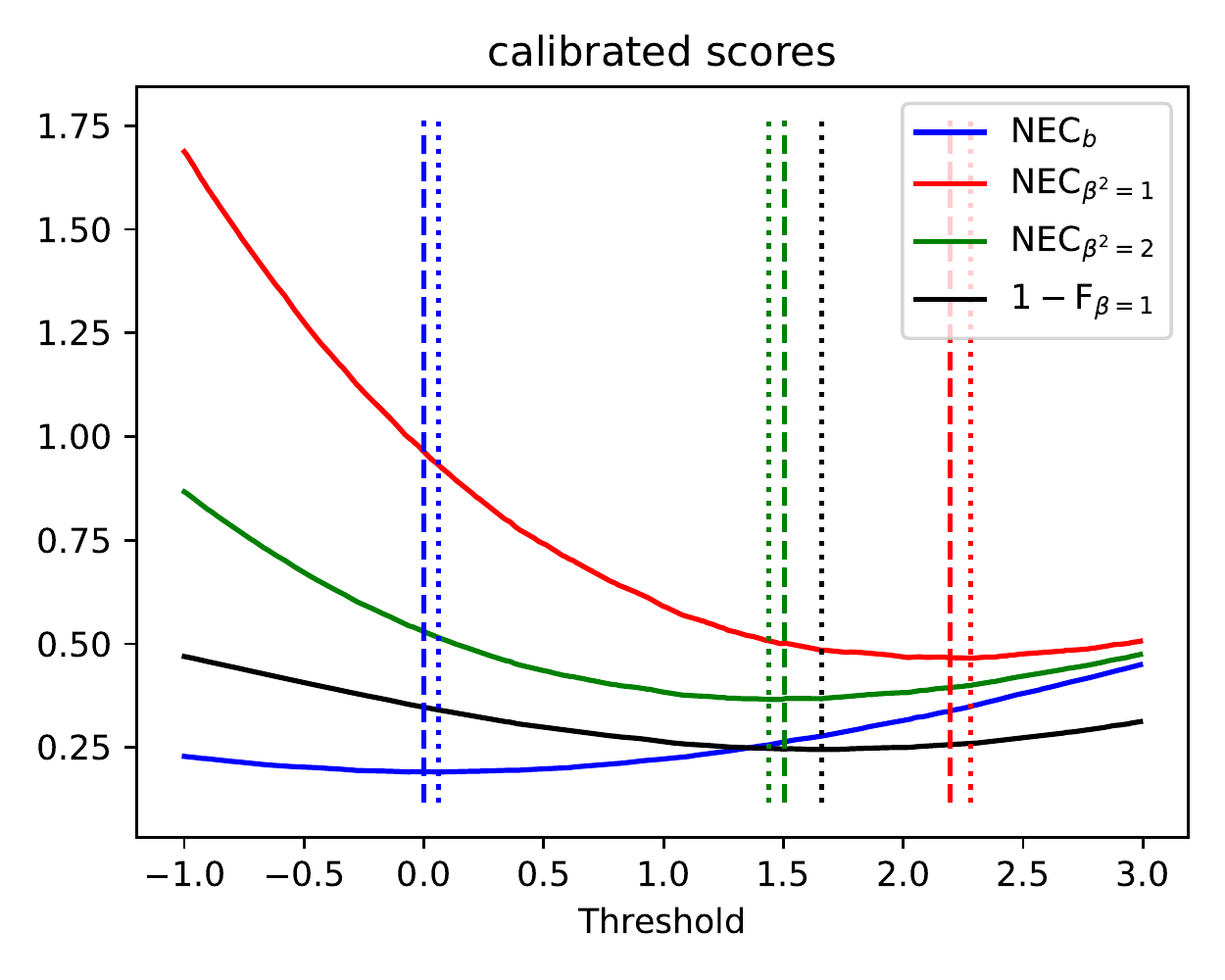}
\caption{Three NEC metrics and 1-$\FSO$ as a function of the threshold for miscalibrated (LR-mc1) and calibrated (LR-cal) scores on a simulated dataset with binary labels. Dotted vertical lines correspond to the optimal threshold for the metric with the same colored lines. Dashed vertical lines indicate the Bayes threshold corresponding to each NEC metric.}
\label{fig:metrics_vs^{(t)}hr}
\end{figure*}

\subsubsection{EC with an Abstain Decision for Binary Classification}

In this section we use the dataset of LR-cal and LR-mc1 scores, as in the previous section, to show the effect of including an abstention decision. Hence, unlike the prior example where the set of decisions $\mathcal D$ was equal to the set of classes $\mathcal H =\{H_1, H_2\}$, here $\mathcal D = \{H_1, H_2, \mathrm{abstain}\}$. We consider the following cost matrix:
\begin{gather}
\mathbf{C} =
  \begin{bmatrix}
   0 & 1 & \alpha \\
   1 & 0 & \alpha \\
   \end{bmatrix}
\end{gather}
The first two columns correspond to the first two decisions: $H_1$ and $H_2$. These decisions are penalized by a cost of 0 if the true class was equal to the decision or by a cost of 1 if it was not. The third column corresponds to the ``abstain" decision where the system does not choose either class. In this case, the penalization is $\alpha$ for both classes (i.e., we consider an abstention to be equally costly for samples of both classes). 

For this cost matrix, the quantity we need to minimize to obtain the best decision for a certain sample $x$ is:
\begin{eqnarray}
 \sum_{i=1}^{K} c_{ij} P(H_i|x) = \begin{cases}
    P(H_2|x), & \text{if } d=H_1  \\
    P(H_1|x), & \text{if } d=H_2  \\
    \alpha  P(H_1|x) + \alpha P(H_2|x) = \alpha, & \text{if } d=\mathrm{abstain}.  \\
\end{cases}  
\end{eqnarray}
This expression corresponds to the sum in \eref{eq:bayes_decisions}, with $d$ taking one of the three values in $\mathcal D$ defined above and $C(H_i, D_j)=c_{ij}$ given by the $(i,j)$ entry in the matrix $\mathbf{C}$ above.
The system will abstain when $\alpha$ is smaller than either of the two posteriors. Since the two posteriors sum to 1, this happens when the maximum posterior is smaller than $1-\alpha$. That is, the Bayes decision will be given by:
\begin{eqnarray}
 d_B = \begin{cases}
    H_1, & \text{if } P(H_2|x) < \min(P(H_1|x), \alpha)  \\
    H_2, & \text{if } P(H_1|x) < \min(P(H_2|x), \alpha)  \\
    \mathrm{abstain}, & \text{if } \max(P(H_1|x), P(H_2|x)) < 1-\alpha  \\
\end{cases}  
\end{eqnarray}
Since $\max(P(H_1|x), P(H_2|x))$ can never be smaller than 0.5, any $\alpha$ larger than 0.5 will result in zero abstentions. For any $\alpha<0.5$, abstentions will happen when the maximum posterior is smaller than $1-\alpha$, that is, when the system is not certain enough about either of the two classes. As $\alpha$ decreases, the number of samples for which that happens increases.

Table \ref{tab:ec_with_abstention} shows the EC for Bayes decisions made with misscalibrated and calibrated LRs for a series of $\alpha$ values. We can see that, in this example, the EC and NEC are significantly lower if we use calibrated scores than if we use the misscalibrated scores, illustrating the importance of ensuring that the system outputs well-calibrated scores when using Bayes decisions.
The table also shows the percentage of samples for which an abstention decision was made in each case, which, as explained above, grows as $\alpha$ decreases. We can see that, for most values of $\alpha$, for this particular system, the raw scores lead to more abstentions than it is optimal (i.e., than for the calibrated scores). Further, note that the EC values show that, by adding an abstention option, we effectively reduce the overall cost (compare the EC values for $\alpha<0.5$ with the EC value at $\alpha=1$ which is equivalent to not having an abstention decision), since abstentions are only made when they are less costly than selecting one of the two classes. This effect is not observed on the normalized EC because the EC of the naive system that is used for normalization is given by $\min(P_1, P_2, \alpha)$ (in our case, $P_1=1-P_2=0.9$), which is 0.1 for all cases except for the first row where it is 0.01, breaking the monotonic decrease in NEC.

\begin{table}[t]
    \centering
    \begin{tabular}{c|ccc|ccc}
    & \multicolumn{3}{c|}{Miscalibrated Scores} & \multicolumn{3}{c}{Calibrated Scores}\\
$\alpha$ & EC & NEC  & Abs  & EC & NEC  & Abs  \\
\hline
 0.010  &  0.006 &  0.638 &    61.5  &  0.005  &    0.534  &   41.1 \\
 0.100  &  0.030 &  0.299 &    12.3  &  0.025  &    0.249  &   14.1 \\
 0.200  &  0.052 &  0.521 &     6.7  &  0.035  &    0.355  &    8.2 \\
 0.400  &  0.076 &  0.756 &     2.0  &  0.046  &    0.456  &    2.3 \\
 0.600  &  0.079 &  0.786 &     0.0  &  0.047  &    0.467  &    0.0 \\
 1.000  &  0.079 &  0.786 &     0.0  &  0.047  &    0.467  &    0.0 \\
    \end{tabular}
    \caption{EC and normalized EC (NEC) for a cost matrix for two classes with an abstention decision with cost of $\alpha$ for both classes. Columns correspond to the EC, NEC, and number of abstentions made with Bayes decisions using the misscalibrated (LR-mc1) and the calibrated (LR-cal) scores. For $\alpha>0.5$, when the abstention decision is never selected, the EC coincides with the total error rate, which is equal to one minus the accuracy.}
    \label{tab:ec_with_abstention}
\end{table}

\subsubsection{PSRs and Calibration Analysis}
\label{sec:examples_cal}

In this section we compare various PSRs for the multi-class ($K=10$) simulated dataset. We include two ECs, one with a square cost matrix and another with an abstain decision. In both cases we take $c_{ij}=0$ for $i=j$, and $c_{ij}=1$ if $i\neq j$. For the cost with an abstention decision, we set the cost to $\alpha=0.1$ for all classes. We also show normalized cross-entropy and normalized Brier score. We use the priors in the test data to compute all metrics.

We compare these metrics using 6 different sets of posteriors described above to make the Bayes decisions (Datap/Mismp-cal/mc1/mc2). Further, we create calibrated versions (affcal) for the misscalibrated posteriors by applying a restricted affine transformation of the form given by \eref{eq:affcal} trained by minimizing cross-entropy. The calibration is done using cross-validation on the test data (\eref{eq:crossent_ave}).  Finally, we show results for a second calibration transformation (temcal) where we only use the scale parameter and set the shift to 0. This is the standard temperature scaling approach used in most recent calibration works \cite{guo:17}. 

Table \ref{tab:ec_multiclass} shows the results for all 6 scores and their two calibrated versions. The results for each metric are organized in three columns, cal, mc1 and mc2, to facilitate analysis. The first block of results corresponds to posteriors obtained with the data priors (Datap) while the second block of results is obtained with the  posteriors computed with mismatched priors (Mism). 

Comparing the original posteriors without a calibration transformation (the first row in each block) across cal, mc1 anc mc2 columns, we can see the effect of miscalibration on each metric. We see that mc2 does not affect the NEC defined with the 0-1 costs for which Bayes decisions are given by the class with maximum posterior, since the scaling performed to obtain those scores does not change the ranking. Note that this NEC is a normalized version of the total error. The normalization factor for this data is given by 0.1, obtained by plugging the priors and costs for this example in \eref{eq:ecn}. Hence, for example, the total error for the Datap-cal system is 0.027, while for the Mismp-cal system it is 0.111. These correspond to 97.3\% and 88.9\% accuracy, respectively. The advantage of using the NEC instead of the accuracy is that it immediately highlights cases that are worse than the naive baseline, as is the case for the Mismp-cal which has a NEC above 1.0.

Comparing results across rows, we can see that for the mc1 miscalibrated posteriors and for the posteriors obtained with mismatched priors (Mismp block), temperature scaling does not solve the calibration problem. This is because the misscalibration in these cases is not restricted to a scaling factor in the log-posteriors, which is the assumption made by this approach. On the other hand, the affine calibration transformation leads to perfect calibration. That is, results with affcal are the same as for the Datap row on the cal column which correspond to perfectly calibrated likelihoods used to compute posteriors with perfectly matched priors. For mc2 with Datap posteriors, temperature scaling works as well as the affine transform since the misscalibration in those scores is given (by construction) by a scaling factor in the log posteriors. 

\begin{table}[t]
%\footnotesize
    \centering
    \begin{tabular}{l|ccc|ccc|ccc|ccc}
    & \multicolumn{3}{c|}{NEC} & \multicolumn{3}{c|}{NEC with abs} & \multicolumn{3}{c|}{Cross-entropy} & \multicolumn{3}{c}{Brier Score} \\
   & cal & mc1 & mc2 & cal & mc1 & mc2 & cal & mc1 & mc2 & cal & mc1 & mc2  \\
\hline
Datap          & 0.25  & 0.29  & 0.25  & 0.14  & 0.17  & 0.93  & 0.13  & 0.17  & 0.57  & 0.21  & 0.26  & 0.70  \\
Datap-temcal   & 0.25  & 0.29  & 0.25  & 0.14  & 0.17  & 0.14  & 0.13  & 0.17  & 0.13  & 0.21  & 0.26  & 0.21  \\
Datap-affcal   & 0.25  & 0.25  & 0.25  & 0.14  & 0.14  & 0.14  & 0.13  & 0.13  & 0.13  & 0.21  & 0.21  & 0.21  \\
\hline
Mismp          & 1.11  & 0.70  & 1.11  & 0.52  & 0.61  & 1.00  & 0.50  & 0.48  & 0.99  & 0.86  & 0.70  & 1.58  \\
Mismp-temcal   & 1.11  & 0.70  & 1.11  & 0.51  & 0.39  & 0.51  & 0.50  & 0.40  & 0.50  & 0.86  & 0.58  & 0.86  \\
Mismp-affcal   & 0.25  & 0.25  & 0.25  & 0.14  & 0.14  & 0.14  & 0.13  & 0.13  & 0.13  & 0.21  & 0.21  & 0.21  \\
    \end{tabular}
    \caption{Various PSRs for 18 sets of posteriors obtained from calibrated likelihoods (cal) or miscalibrated likelihoods (mc1) using either the priors in the test data (Datap) or mismatched priors (Mismp). For each set of priors, a second set of miscalibrated scores (mc2) is obtained by scaling and renormalizing the corresponding cal posteriors. Finally, for each of those scores, two calibrated versions are obtained with an affine transform (affcal) and with temperature scaling (temcal). The accuracy for each system can be obtained as $1-0.1\ECn$.}
    \label{tab:ec_multiclass}
\end{table}

Table \ref{tab:ec_multiclass_calloss} shows calibration loss measured with each of the two strict EPSRs (cross-entropy and Brier score) using \eref{eq:relcalloss}, where $\mathrm{EPSRraw}$ is the EPSR in the corresponding cell in Table \ref{tab:ec_multiclass} and $\mathrm{EPSRemin}$ is the EPSR value within the same block and column after affine calibration (affcal). This calibration loss tells us what percentage of the PSR in Table \ref{tab:ec_multiclass} is due to misscalibration, assuming that the affine calibration is doing the best possible job at calibrating the scores. In this specific example, this is the case, since the misscalibration is perfectly modelled by an affine transformation by construction. When this is not the case, other calibration transforms from the large variety proposed in the literature could be explored (see Section \ref{sec:calibration_loss}). For comparison, the table also includes the ECEmc values computed using 15 bins, which is a standard value for this parameter.

The table shows that, in these examples, cross-entropy and Brier score lead to similar conclusions about the degree of miscalibration of each system. On the other hand, ECEmc fails to detect the calibration problem introduced by using mismatched priors to compute the posteriors (the Mismp-cal system). This is because the ECEmc evaluates only the confidences, i.e., the posterior for the decision made by the system. The Mismp-cal system makes only 10\% of the decisions differently than the Datap-cal system. For the samples for which the decision is the same, the average difference in posterior is 0.07 causing a negligible increase in the ECEmc. For the samples for which the decision does change, the difference in confidence between the two systems is 0.25. Since this happens for relatively few samples, the ECE is not greatly affected by this difference. On the other hand, the calibration loss based on Brier score and cross-entropy clearly show the severity of the misscalibration produced by using mismatched priors to compute the posteriors. This is because these metrics evaluate the full posterior vector and these vectors are wrong for every single sample. This is, we believe, a very compelling example against the use of the ECEmc for assessing calibration performance. 

\begin{table}[t]
%\footnotesize
    \centering
    \begin{tabular}{l|ccc|ccc|ccc}
    & \multicolumn{3}{p{2.3cm}|}{\centering Cal-loss with Cross-entropy} & \multicolumn{3}{p{2.3cm}|}{\centering Cal-loss with Brier Score} & \multicolumn{3}{c}{ECEmc}  \\
   & cal & mc1 & mc2 & cal & mc1 & mc2 & cal & mc1 & mc2  \\
\hline
Datap          &    0  &   23  &   77  &    0  &   20  &   71  &    0  &    2  &   22  \\
Datap-temcal   &    0  &   23  &    0  &    0  &   20  &   0  &    0  &    2  &    0  \\
Datap-affcal   &    0  &    0  &    0  &    0  &    0  &    0  &    0  &    0  &    0  \\
\hline
Mismp          &   74  &   73  &   87  &   76  &   71  &   87  &    2  &    9  &   28  \\
Mismp-temcal   &   74  &   68  &   74  &   76  &   65  &   76  &    1  &    1  &    1  \\
Mismp-affcal   &    0  &    0  &    0  &    0  &    0  &    0  &    0  &    0  &    0  \\
    \end{tabular}
    \caption{Calibration loss (in percentages) for the systems in Table \ref{tab:ec_multiclass} measured using \eref{eq:relcalloss} using Cross-entropy and Brier score as PSRs. ECE values are also included for comparison.}
    \label{tab:ec_multiclass_calloss}
\end{table}

\subsection{Discussion on Metrics for System Scores}

In this section we have discussed various metrics for assessing the performance of a classifier's scores, before making categorical decisions. In particular, we focus on classifiers for which the scores are posterior probabilities. Such systems have the important benefits of producing interpretable outputs and of enabling the use of Bayes decision theory for making optimal categorical decisions.

While PSRs were proposed decades ago to measure the quality of posterior probabilities, they are surprisingly underused as evaluation metrics in current machine learning literature. Instead, the ECE or some of its variants stand as the main metric used in works where posterior probabilities are required. As we argue above, the ECE, being a calibration-only metric, does not appropriately assess the quality of posteriors since it only reflects one aspect of the system performance. Hence, it should not be used to make decisions about the quality or interpretability of posteriors. Calibration metrics like the ECE are very useful during development, though, to allow us to analyze whether adding a calibration stage to our system would improve the quality of its posteriors. Yet, even for that purpose, ECE is not an ideal metric for various reasons including the fact that its absolute value cannot be directly interpreted, that for the multi-class case it only evaluates the quality of the confidences instead of that of the full vector of posteriors, and that it relies on histogram binning which is a rather fragile calibration approach. 

In this section we have discussed why the expected value of a PSR is a principled metric for evaluating the quality of a system's posteriors. Expected PSRs can also be used to estimate the calibration quality of the posteriors by computing the minimum value of the expected PSR that can be obtained after calibration. The difference between the actual and the minimum value of the expected PSR, called calibration loss, can be used as a more principled, robust and interpretable replacement for the ECE. 

While expected PSRs are calibration-sensitive metrics, other metrics like the AUC and the EER measure only the discrimination performance of the system. Their values are immune to any invertible transformation of the scores and, hence, are not sensible to calibration problems. They are, in this sense, comparable to the expected PSR obtained after calibration, EPSRemin. Being calibration-insensitive metrics, they ignore the problem of making categorical decisions, implicitly assuming that we will find a way to make those decisions optimally. If the goal of the evaluation is to asses the actual performance that the hard decisions will have when the system is deployed, a calibration-sensitive metric like the expected value of a PSR (e.g., the EC for Bayes decisions, the cross-entropy, or the Brier score) should be used for evaluation instead of discrimination-only metrics like the AUC or the EER.

\section{Conclusions}

This paper presented and compared various metrics commonly used in the literature for training and evaluating classification systems. We first described metrics that evaluate the quality of hard decisions, including expected cost (EC), standard and balanced total error (which are special cases of the EC), the standard and balanced accuracy (which are one minus the corresponding total error), F-beta score and others. Then, we described metrics designed to evaluate the quality of the system's scores, including the area under the ROC curve, the equal error rate, and the expected value of proper scoring rules (PSRs), of which the cross-entropy is one special case. PSRs assume that the system's output will be used as posterior probabilities to make optimal Bayes decisions. For this reason, they can be used to evaluate the quality of a system's calibration, providing a more elegant solution to this problem than the widely used expected calibration error (ECE). 

While the EC and PSRs have been discussed in the statistical learning literature for decades, they have not been widely adopted by the machine learning community.
The main goals of this paper were: (1) to highlight the superiority of the EC over other standard and more widely used metrics for hard decision like the F-beta score, (2) to describe the desirable properties of the PSRs for evaluating classification system outputs, and (3) to argue that calibration loss computed using PSRs offers many advantages over the widely used ECE. We gave both theoretical and practical arguments to support these claims. 
The paper is accompanied by an open source repository which provides code for the computation of the various metrics in this paper. All plots and tables in this paper can be replicated using the repository. We hope that this document along with the provided code will facilitate the wider adoption of these metrics that offer a principled, elegant, and general solution to the evaluation of classification systems.

\bibliographystyle{IEEEbib}
\bibliography{all-short.bib}

\end{document}